\documentclass[11pt]{article}

\usepackage[preprint]{acl}

\usepackage{times}
\usepackage{latexsym}
\usepackage{enumitem}
\usepackage{float}

\usepackage[T1]{fontenc}

\usepackage[utf8]{inputenc}
\DeclareUnicodeCharacter{05F3}{'}

\usepackage{microtype}

\usepackage{inconsolata}

\usepackage{graphicx}
\usepackage{amsmath}
\usepackage[most]{tcolorbox}
\usepackage{amssymb}
\usepackage{tabularx}
\usepackage{booktabs}
\usepackage{tcolorbox}
\usepackage{subcaption}
\usepackage{tikz}
\usetikzlibrary{arrows.meta, positioning}

\title{PEMAND: Persona-Enriched Multi-Agent Negotiation for Household Decision-Making}

\author{
 \textbf{Yuran Sun\textsuperscript{1}\footnotemark[1]},
 \textbf{Mustafa Sameen\textsuperscript{1}\footnotemark[1]},
 \textbf{Yaotian Zhang\textsuperscript{1}},
 \textbf{Rongguan Gu\textsuperscript{1}},
\\
 \textbf{Mrunal Vibhute\textsuperscript{1}},
 \textbf{Chia-yu Wu\textsuperscript{1}},
 \textbf{Yuanyuan Lei\textsuperscript{1}},
 \textbf{Xilei Zhao\textsuperscript{1}}
\\
 \textsuperscript{1}University of Florida
\\
 \small{
   \textbf{Correspondence:}
   \href{mailto:yuransun@ufl.edu}{yuransun@ufl.edu}
 }
}

\begin{document}

\maketitle
\footnotetext[1]{Equal Contribution.}
\begin{abstract}
Modeling household-level decisions is central to many real-world applications, including trip planning, residential mobility and migration, disaster management, etc. Existing studies primarily rely on classical machine learning models with limited predictive capacity, while recent LLM-based approaches have yet to incorporate behavioral theory or intra-household interaction dynamics, both of which are essential for modeling realistic household decisions. To address these limitations, we propose Persona-Enriched Multi-Agent Negotiation for household Decision-making (PEMAND), a novel LLM-based framework that integrates behavioral theory into individualized, household-aware persona modeling and simulates household-level decision-making through structured multi-agent negotiation. Specifically, PEMAND transforms static sociodemographic attributes into coherent narrative profiles that explicitly encode household-level attitudes, subjective norms, and perceived behavioral controls, following our proposed Household-Aware Chain-of-Planned-Behavior (HA-CoPB) framework. Building on these theory-grounded personas, PEMAND captures real-world household decision negotiation via a structured two-phase multi-agent conversation framework with a novel persona-alignment control mechanism. Evaluated on national and regional household decision datasets across two major domains, including travel behavior and residential mobility, PEMAND consistently outperforms state-of-the-art benchmarks.
\end{abstract}

\section{Introduction}
Household decision-making is fundamental in real-world applications such as transportation planning \cite{talpur2023computing}, urban development \cite{wang2021factors}, emergency management \cite{cova2024destination}, and retail \cite{erasmus2014consumers}. However, modeling these decisions remains challenging for AI and machine learning (AI/ML), as household decision-making emerges from dynamic interactions among members’ preferences, roles, constraints, and multi-stage cognitive processes \cite{ajzen1991theory, wang2007cognitive, golob1997model, yang2016research, leyer2025understanding}. This creates a critical need for novel methodological frameworks that can capture such complexities, to improve predictive accuracy and support policy and planning across diverse decision domains \cite{acosta2020does}.

Prior machine learning approaches for modeling household decisions \cite{sun2024predicting, zhang2019household, xue2022adopting, firoozzare2024understanding, amarasinghe2022understanding} often treat the household as a monolithic unit or a simple aggregation of individuals, failing to capture complex intra-household interactions \cite{dong2008studying, anand2026machine, mu2026collective}.

While large language models (LLMs) can leverage unstructured information and produce flexible representations of household contexts and decision factors \cite{xiao2025evaluating, chen2025perceptions, lim2026can}, and multi-agent systems make it possible to simulate intra-household interaction, coordination, and negotiation dynamics \cite{abdelnabi2023llm, abdelnabi2024cooperation}, they face three critical limitations when applied to decision prediction. \textit{First}, existing LLM frameworks lack explicit grounding in multi-stage cognitive theories \cite{ajzen1991theory, wang2007cognitive}, and even when enriched with sociodemographic and contextual information, they may partially capture the underlying decision-making mechanisms without behavioral guidance \cite{chen2025perceptions, tjuatja2024llms}. \textit{Second}, they rely on generic dialogue protocols rather than structured communication that reflects real-world household deliberation processes, where members exchange preferences, coordinate constraints, and resolve conflicts before reaching a collective outcome \cite{golob1997model, kong2025survey}. \textit{Third}, although LLMs provide expressive representations and strong reasoning capabilities, they remain prone to hallucinations \cite{bang2025hallulens}, invalid or inconsistent outputs \cite{ahn2025prompt}, and misalignment with observed human behavior \cite{lei2024fairmindsim, wang2023aligning}.

To address these limitations, we propose \textbf{Persona-Enriched Multi-Agent Negotiation for household Decision-Making (PEMAND)}, a novel two-stage LLM-based framework that bridges theory-grounded cognitive modeling with collective household negotiation. Our approach simulates household decisions by assigning each LLM agent the role of a household member (e.g., mother, father, child) and guiding the decision process through two phases. First, we synthesize factually-grounded personas using a heuristic anchoring mechanism \cite{tversky1974judgment} to ensure agents reflect specific sociodemographic distributions rather than generic stereotypes. Specifically, we extend the Chain-of-Planned-Behavior (CoPB) framework \cite{Shao2024} into a Household-Aware CoPB (HA-CoPB) formulation grounded in the Theory of Planned Behavior (TPB) \cite{ajzen1991theory}, a well-established framework for modeling human decision-making. HA-CoPB adapts TPB components to the household setting by representing Perceived Behavioral Control (PBC) as shared resource constraints, such as vehicle availability, and Subjective Norms (SN) as concrete intra-household dependencies. To improve persona reliability, we further introduce an Alignment via Refinement (AvR) pattern \cite{madaan2023self}, where a review agent critiques the generated reasoning against historical priors \cite{tversky1974judgment} and household constraints, prompting the primary agent to iteratively revise its behavioral profile. In the second stage, these agents engage in a structured negotiation protocol to resolve conflicting needs and converge on a realistic, consensus-based household decision. This process consists of two phases: (1) a parallel proposal phase that captures heterogeneity in individual intentions, and (2) a consensus refinement phase that models intra-household interaction and collective decision-making. We further introduce an alignment control mechanism consisting of a human-calibrated LLM-judge moderator, persona supervised fine-tuning (SFT), and negotiation reinforcement self-training (ReST). This mechanism encourages each agent to adhere to its assigned HA-CoPB, interact validly with other household agents, and produce decisions that better align with observed real-world behavior.

Through evaluations from two distinct domains, including trip planning and relocation decisions, using national and regional scale survey datasets, we compare PEMAND with a suite of state-of-the-art baselines and show that PEMAND consistently achieves superior performance across datasets. 

Our primary contributions include:
\begin{itemize}[leftmargin=*, nosep]
    \item \textbf{A Novel Multi-stage Methodological Framework for Household Decision Modeling.} We introduce a new multi-stage formulation for household decision prediction that treats household decisions as collectively negotiated outcomes rather than independent individual predictions. Unlike prior LLM-based decision models based on static personas or direct prompting, PEMAND unifies household-aware persona construction, role-specific proposal generation, constraint-aware negotiation, and alignment control in a single inference pipeline.
    
    \item \textbf{Behavioral Theory-Guided Persona Inference.} We introduce a theory-grounded multi-agent architecture that embeds TPB-based cognitive reasoning into household simulation. By adapting TPB to shared household constraints and enforcing a hierarchical priority mechanism, PEMAND anchors each agent’s intentions in constrained HA-CoPB profiles before negotiation, reducing unconstrained hallucination beyond static persona-based prompting.

    \item \textbf{Structured Multi-agent Negotiation with Alignment Control.} We introduce a structured household negotiation and alignment mechanism that models collective decisions as a controlled deliberation process rather than generic multi-agent dialogue. The alignment module uses human calibration to ground the LLM judge in expert-validated behavioral criteria, providing an external alignment signal beyond model self-assessment.
\end{itemize}

\section{Related Work}
\textbf{Modeling Household Decision.} Household decision-making has been widely studied across multiple domains, including transportation planning \cite{anand2026machine, anand2025comparative}, residential mobility \cite{wang2021factors, clark2017decisions}, emergency management \cite{sun2024predicting, cova2024destination}, and retail marketing \cite{erasmus2014consumers, inan2025ethical}. Existing studies have employed a range of modeling approaches to analyze and predict household decisions, including statistical models \cite{clark2017decisions, al2025determinants, kuligowski2022modeling, qawasmeh2024estimation}, machine learning models \cite{anand2026machine, sun2024predicting, naseralavi2025machine, qawasmeh2024estimation}, economic models \cite{buchmann2025good}, and recently, LLM models \cite{chetty2025llm, pijpinsights}. Existing models often treat households as single decision-making units or aggregate individual choices, overlooking intra-household interaction dynamics \cite{mu2026collective, dong2008studying, anand2026machine}. Even theory-grounded models \cite{kuligowski2022modeling, sun2024investigating} still struggle to capture realistic multi-stage decision processes.

\noindent\textbf{Persona Generation for Decision Modeling.} Persona generation for human decision modeling has shifted from static demographic profiles to richer, psychologically grounded LLM-based representations. Recent work uses LLMs to infer latent cognitive antecedents \cite{Shao2024, sameen2025sapa}, synthesize daily activity schedules from aggregate data \cite{amin2025generating}, and forecast long-term life-course decisions such as residential mobility from textual trajectories and panel datasets \cite{stark2025using, gao2026how}.

Despite this progress, key gaps remain for joint household-level prediction. Standard zero-shot initialization can suffer from population misalignment and persona degradation during multi-agent dialogue \cite{Hu2025, frisch2024llm}. Existing isolated generation methods also miss intra-household entanglement, including shared resources, relational constraints, and interpersonal obligations \cite{liu2025humanmobilitymodelinghousehold}. Moreover, independent personas often lack the communicative agency and constraint enforcement needed for structured negotiation, leading to implausible individual outcomes and weak collective consensus.

\noindent\textbf{LLM-based Multi-agent Conversation.} Recent work shows that communication among LLM-based agents can improve collective task performance \cite{zhang2024cut, yan2025beyond, wu2024autogen}. A common paradigm assigns agents explicit roles or personas, enabling structured collaboration and more realistic role-playing interactions \cite{abbasiantaeb2024let, wu2024autogen, li2023camel, feng2025reframe}. Richer persona specifications, including background, emotional state, and psychological traits, further improve behavioral alignment in dialogue \cite{shao2023character, yang2025psyplay, wu2025personas}. However, most role-playing multi-agent frameworks still rely on prompt-based persona conditioning without explicit mechanisms for sustained alignment with real-world interaction dynamics, limiting their use in behaviorally grounded decision-making.

\section{Methodology}
\begin{figure*}[t]
    \centering
\includegraphics[width=0.95\linewidth, trim=0cm 0.5cm 0cm 0cm, clip]{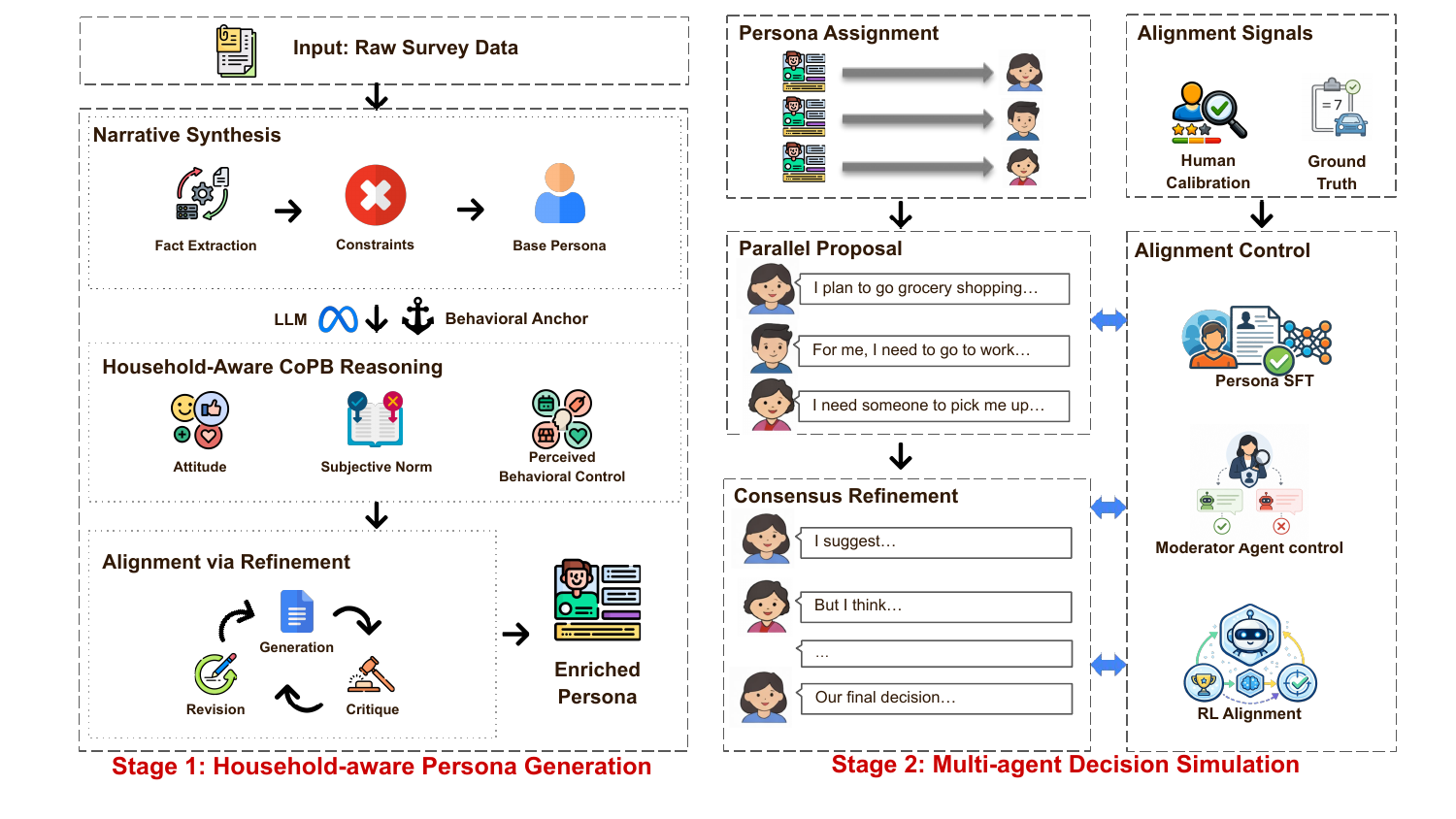}
    \caption{Methodological Framework for PEMAND.}
    \label{fig:method_framework}
    \vspace{-0.15in}
\end{figure*}

\vspace{-0.1in}
We propose a two-stage LLM-based framework for household decision prediction. Rather than directly mapping household attributes to outcomes, our framework models the intermediate processes through which household decisions are formed, including individual reasoning and intra-household negotiation. Given a household $\mathcal{H}$ with $N$ members $\{$$m_1$, $m_2$, $\dots$, $m_N$$\}$, along with their demographic and contextual attributes, the goal is to predict the final household-level decision, such as trip count or relocation choice. We formulate this task as a multi-stage process in which each member first generates a persona-grounded individual proposal, after which members engage in structured negotiation to reach a collective decision. Figure~\ref{fig:method_framework} provides an overview of the proposed framework.

\subsection{Behavioral Theory-Guided Persona Generation}
Our persona generation module bridges static demographic data and dynamic behavioral reasoning by integrating deterministic fact extraction with constrained generative reasoning. This approach ensures that the synthesized narrative is factually accurate and structurally consistent with the overall household composition. More importantly, the framework is domain-agnostic and can be instantiated for different household-level decision tasks.

\subsubsection{Narrative Synthesis and Behavioral Anchor}
\textbf{Deterministic Fact Extraction.} We first employ a deterministic \textit{Translation Map} ($\Phi: \mathcal{X} \rightarrow \mathcal{F}$) that converts encoded survey responses into atomic natural language facts (e.g., $\Phi(\texttt{EDUC}=5)$ maps to ``\textit{I have a graduate or professional degree.}''). This step utilizes a non-generative approach to ensure data fidelity. We also apply a \textit{Neutralization Filter} at this stage to map ambiguous non-response codes to generic descriptors, preventing downstream hallucination of unfounded relationship dynamics.

\noindent\textbf{Narrative Synthesis with Faithfulness Constraints.} We then feed the atomic facts $\mathcal{F}$ into an instruction-tuned LLM to synthesize a first-person narrative $\mathcal{P}$. To ensure data faithfulness and mitigate sociodemographic hallucination \cite{ji2023survey}, the system prompt imposes strict \textit{Consistency Constraints} and task-specific \textit{Vocabulary Restrictions}. This guarantees a clinically neutral but cognitively coherent persona profile that serves as the agent's implanted memory, aligned with its ground-truth attributes (detailed in Appendix~\ref{app:prompts}).

\textbf{Behavioral Anchor (Historical Prior).} To ground the LLM's reasoning in statistical reality and prevent data leakage \cite{hyndman2018forecasting}, we employ a historical \textbf{Behavioral Anchor} mechanism. We guide the LLM using a population-level reference point derived strictly from past survey cycles. This operationalizes the \textit{Anchoring and Adjustment Heuristic} \cite{tversky1974judgment} within a Bayesian framework \cite{washington2020statistical}, where the model treats the historical anchor as a prior and updates it based on specific persona constraints.

\subsubsection{Household-Aware CoPB (Reasoning Engine)}
Our \textit{Household-Aware CoPB (HA-CoPB)} framework models the agent's cognitive process as a generative transformation, $f_{\text{HA-CoPB}}$, mapping the static persona to an \textbf{Enriched Persona} ($\mathcal{P}'$):
\begin{equation}
\mathcal{P}' = f_{\text{HA-CoPB}}(\mathcal{P}, \mathcal{H}_{ctx} \mid y_{\text{prior}}^{t-\text{lag}})
\end{equation}
where $\mathcal{H}_{ctx}$ represents the shared household context (e.g., resource bottlenecks, household member attributes). This formulation is grounded in the \textbf{Theory of Planned Behavior (TPB)} \cite{ajzen1991theory}, a well-established theory of decision-making, and adapts its core psychological constructs to the household level (Appendix~\ref{app:tpb_diagram}):
\begin{equation}
\mathcal{P}' = \mathcal{P} \cup \{ \mathcal{A}, \mathcal{SN}, \mathcal{PBC} \}
\end{equation}
These constructs capture individual intrinsic utility ($\mathcal{A}$), intra-household obligations ($\mathcal{SN}$), and shared resource barriers ($\mathcal{PBC}$). To prevent hallucination, generation follows a strict Hierarchical Priority Mechanism ($\mathcal{PBC} \succ \mathcal{SN} \succ \mathcal{A}$), mandating that physical resource limits override social obligations and individual desires.

\subsubsection{Alignment via Refinement (AvR) Mechanism}
To synthesize the individual reasoning generated by the HA-CoPB framework into a robust psychological profile, we implement an \textbf{Alignment via Refinement (AvR)} architecture. Across all domains, the persona is finalized through an iterative dual-agent cycle. A primary reasoning agent (the Actor) leverages the HA-CoPB template to generate an initial persona narrative based on demographics and behavioral anchors. Subsequently, a secondary review agent (the Reviewer) evaluates this rationale, contrasting it against historical population anchors and structural household constraints (e.g., resource deficiencies). The Reviewer provides explicit critique, prompting the Actor to revise and calibrate its reasoning. This generation-critique-revision loop ensures that the resulting \textbf{Enriched Persona} is both theoretically grounded and behaviorally realistic, preventing unconstrained hallucination before any downstream decision-making occurs.

\subsection{LLM-based Multi-agent Conversation} 
\subsubsection{Multi-agent Component}
We design a simulation framework that captures realistic intra-household discussions for household decision-making, while incorporating additional control mechanisms to improve LLM alignment. The system comprises $N$+1 agents, {$A_1$, $A_2$, $\cdots$, $A_{N+1}$}, where $N$ corresponds to the number of household members. The first $N$ agents represent household members with distinct roles, sociodemographic attributes, and personas, while $A_{N+1}$ serves as a moderator that enforces discussion-level constraints and supports coherent, aligned interaction dynamics.

\noindent \textbf{Household Member Agent.} Household member agents simulate participants in household discussion and collective decision-making. Each agent represents a distinct household role with corresponding behavioral patterns, preferences, and constraints. To link generated personas to negotiation behavior, every member agent is assigned HA-CoPB persona-specific attributes that guide its responses and interactions, as described below:
\begin{equation}
    P_i = \{role_i, \mathcal{P}^{'}_i, \mathcal{H}_{ctx_i}\}
\end{equation}
where $role_i$ specifies the agent’s role within the household (e.g., parent or child) and indicates whether the agent serves as the household lead.

\noindent \textbf{Moderator.} The moderator is a non-member, human-calibrated LLM judge responsible for intra-conversation control (Appendix ~\ref{app: moderator}). To ensure its evaluations reflect real-world decision standards, it is rigorously calibrated against human annotators (calibration metrics are detailed in Appendix~\ref{app:moderator-calibration}). It monitors the discussion using member attributes, shared household resources, and dialogue context. After each utterance, it checks whether the response is persona-consistent, constraint-compliant, discussion-relevant, and supported by a justified decision or proposal.

\subsubsection{Multi-phase Multi-agent Household Discussion Framework}
Prior work views household decision-making as shaped by both individual behaviors and household-level characteristics. Members form initial intentions from heterogeneous preferences and behavioral tendencies \cite{wang2024large, li2024more}, while final household travel decisions emerge through negotiation, compromise, and adaptation to shared resource constraints \cite{golob1997model, naseralavi2025machine}.

We propose a two-phase interaction protocol for realistic household-level decision-making that captures both individual intention formation and intra-household negotiation. In the \textbf{parallel proposal phase}, members independently express their initial intentions; in the \textbf{consensus refinement phase}, agents iteratively deliberate and revise their proposals to reach a collective household decision.

\noindent\textbf{Parallel Proposal.}
In the first phase, each household member agent independently generates an initial decision proposal conditioned on its persona profile (Appendix~\ref{sub: parallel}). For agent $A_i$, the initial vote $v_i^{(0)}$ is sampled as:
\begin{equation}
v_i^{(0)} \sim \pi_{\theta_i}(\cdot \mid P_i).
\end{equation}
Parallel generation elicits heterogeneous intentions while avoiding turn-order and anchoring biases. The individual votes are then aggregated into an initial household-level decision $\hat{y}^{(0)}$:
\begin{equation}
\hat{y}^{(0)} = G(v_1^{(0)}, v_2^{(0)}, \cdots, v_N^{(0)}),
\end{equation}
where $G(\cdot)$ denotes a task-specific aggregation operator, such as summation for numerical decisions or majority voting for binary decisions. This aggregate serves as the starting point for the subsequent household conversation.

\noindent\textbf{Consensus Refinement.}
In the consensus refinement phase, member agents deliberate over the household decision using a round-robin protocol for up to $R_{\max}$ rounds. At round $t$, each agent generates an utterance $u_i^{(t)}$ conditioned on its persona attributes $P_i$, the current household estimate $\hat{y}^{(t-1)}$, compact summaries of other members $C_{-i}$, and the conversation history $h_{t-1}$ (Appendix ~\ref{sub: refinement}):
\begin{equation}
u_i^{(t)} \sim \pi_{\theta_i}(\cdot|P_i, \hat{y_i}^{(t-1)}, C_{-m}, h_{t-1})
\end{equation}
Each utterance may express agreement or disagreement with the current household-level estimate, introduce evidence-grounded revisions to the proposed decision, or negotiate compromises among members’ preferences and constraints.
After each round, we extract each speaker’s preferred decision and apply revision clamps to numerical outputs to reduce sampling-induced fluctuations, while binary decisions remain unchanged. Details of the clamping procedure are provided in Appendix~\ref{app: clamping}.
The discussion ends when all speaker agents agree on a preferred decision or when $R_{\max}$ is reached. If no consensus is reached, the final household decision is computed from the speakers’ final preferences using the median for numerical outputs or majority vote for binary outputs.

\subsubsection{Role-aligned Control Mechanism}
\label{sub: control}
Reliable LLM alignment remains a key challenge in multi-agent household decision modeling. Prior work uses persona specifications to induce human-like, heterogeneous behavior \cite{amin2025generating, wang2024large}, but persona conditioning alone cannot ensure profile consistency or realistic, coherent negotiation across multi-turn interactions \cite{wang2024large}. 

To address these limitations, we introduce two role-aligned control mechanisms: intra-conversation control, where a moderator validates agent behavior during simulation, and out-of-conversation supervision, where role-conditioned SFT and ReST-based trajectory optimization ground member agents before deployment. To ensure reliable moderation, we also design task-specific rubrics and calibrate the LLM judge against human annotations.

\noindent \textbf{Human Calibration.} All LLM judges in the alignment control module are calibrated against human annotations to ensure that they evaluate responses according to expert-validated rubrics rather than model self-assessment. This calibration reduces potential judge-induced bias and improves the reliability of response validation. Calibration details are provided in Appendix~\ref{app:moderator-calibration}.

\noindent \textbf{Intra-conversation Control.} Under intra-conversation control, the moderator acts as a human-calibrated LLM judge that enforces response validity during generation. At each round $t$, it evaluates each utterance $u_i^{(t)}$ along four criteria: persona consistency, constraint adherence, interaction validity, and decision quality, with definitions in Appendix~\ref{app:moderator-calibration}. Utterances that fail to receive full scores on all criteria are discarded and regenerated before entering the conversation history, preventing misaligned or infeasible responses from influencing later interactions.

\noindent \textbf{Out-of-conversation Supervision.}
In the parallel proposal phase, we fine-tune the LLM to generate persona-consistent and constraint-compliant initial decisions. We construct supervision from the training split using an iterative generation-and-filtering procedure: candidate proposals are sampled from the base model and evaluated by the human-calibrated LLM judge for persona consistency and constraint adherence. Only proposals receiving full scores on both criteria are retained as SFT targets, forming a role-specific proposal dataset $\mathcal{D}{i}^{prop} = {(P_i, q_{\text{prop}}, v_i^{(0)\star})}$, where $q_{\text{prop}}$ denotes the proposal prompt.

In the consensus refinement phase, we promote realistic and persona-consistent household negotiation using an inference-time best-of-$N$ ReST trajectory selection strategy. For each training household, we sample $N$ independent discussion trajectories and assign each a weighted reward score:
\begin{equation}
r = -\omega_{GT}\cdot |\hat{y}-y|+\omega_R\cdot \bar{\phi} 
\end{equation}
where $\hat{y}$ is the predicted household decision, $y$ is the ground-truth decision, and $\bar{\phi}$ is the average score across the four PEMAND rubric dimensions from the human-calibrated LLM judge. For each household, we retain the highest-reward trajectory as high-quality self-generated supervision for SFT. At test time, the ReST-tuned policy is deployed under standard inference without access to ground-truth labels.

\section{Experiment}
\subsection{Experiment Setting}
\textbf{Implementation Details} We evaluate PEMAND under homogeneous backbone configurations, using \textbf{Llama-3.1-8B-Instruct} \cite{grattafiori2024llama3herdmodels} end-to-end for persona synthesis, household member agents, and the LLM judge in the main experiments. The judge is calibrated with human annotations and achieves strong reliability: Fleiss’ $\kappa$ \cite{fleiss1971measuring} $=0.620$ and QWK \cite{cohen1968weighted} $=0.698$ for travel, and $\kappa=0.853$ and QWK $=0.851$ for mobility. Full annotation details are provided in Appendix~\ref{app:moderator-calibration}. We prioritize offline open-weight models to support broad adoption, enable SFT, and protect sensitive demographic data. To assess robustness, we also evaluate \textbf{Gemma-3-27b-it} \cite{gemma3_27b_it_hf} and \textbf{GPT-5.4-mini} \cite{openai2026gpt54mini}. Detailed parameter settings, computational costs, latency, and QLoRA fine-tuning hyperparameters are provided in Appendix~\ref{app:implementation}; robustness analyses are reported in Appendices~\ref{app:robustness}.

\subsection{Main Results}

\begin{table*}[t]

  \begin{center}
    \begin{small}
      \begin{sc}
        \resizebox{\textwidth}{!}{
        \begin{tabular}{l cccc cccc cccc}
          \toprule
          & \multicolumn{8}{c}{\textbf{Travel Domain (Regression)}} 
          & \multicolumn{4}{c}{\textbf{Mobility Domain (Classification)}} \\
          \cmidrule(lr){2-9} \cmidrule(lr){10-13}
          & \multicolumn{4}{c}{\textbf{NHTS 2017}} 
          & \multicolumn{4}{c}{\textbf{Puget Sound 2023}} 
          & \multicolumn{4}{c}{\textbf{PSID 2021--2023}} \\
          \cmidrule(lr){2-5} \cmidrule(lr){6-9} \cmidrule(lr){10-13}
          \textbf{Model}
          & MAE$\downarrow$ & RMSE$\downarrow$ & sMAPE$\downarrow$ & $\pm 2$ Acc$\uparrow$
          & MAE$\downarrow$ & RMSE$\downarrow$ & sMAPE$\downarrow$ & $\pm 2$ Acc$\uparrow$ 
          & Acc$\uparrow$ & F1$\uparrow$ & Prec$\uparrow$ & Rec$\uparrow$ \\
          \midrule
          \textit{Traditional ML} \\
          Linear / Logistic Reg.$^\dag$ & 3.34 & 4.52 & 51.00 & 0.40 & 2.83 & 3.86 & 58.54 & 0.58 & 0.69 & 0.55 & 0.47 & 0.67 \\
          Random Forest   & 3.16 & 4.36 & 51.74 & 0.43 & 2.75 & 3.74 & 57.17 & 0.59 & 0.72 & 0.51 & 0.51 & 0.51 \\
          Gradient Boosting   & 3.07 & 4.27 & 51.87 & 0.44 & 2.76 & 3.78 & 57.47 & 0.58 & 0.69 & 0.55 & 0.48 & 0.67 \\
          MLP  & 3.35 & 4.53 & 50.84 & 0.40 & 2.79 & 3.77 & 57.82 & 0.58 & 0.74 & 0.43 & 0.58 & 0.35 \\
          \midrule
          \textit{LLM Baselines} \\
          Demographics-only & 4.06 & 5.59 & 65.29 & 0.44 & 3.31 & 5.10 & 61.75 & 0.54 & 0.59 & 0.40 & 0.35 & 0.49 \\
          Persona  & 3.75 & 5.15 & 63.52 & 0.45 & 2.89 & 3.99 & 59.52 & 0.56 & 0.68 & 0.44 & 0.44 & 0.43 \\
          RAG & 4.32 & 6.18 & 75.25 & 0.41 & 5.00 & 6.60 & 76.49 & 0.35 & 0.61 & 0.48 & 0.39 & 0.63 \\
          Self-consistency & 3.74 & 5.11 & 63.66 & 0.46 & 2.89 & 4.04 & 58.30 & 0.57 & 0.43 & 0.45 & 0.31 & \textbf{0.81} \\
          \midrule
          \textbf{PEMAND}
          & \textbf{2.38} & \textbf{3.93} & \textbf{34.48} & \textbf{0.60}
          & \textbf{1.99} & \textbf{3.25} & \textbf{47.92} & \textbf{0.78} 
          & \textbf{0.79} & \textbf{0.73} & \textbf{0.69} & 0.78 \\
          \bottomrule
        \end{tabular}
        }
      \end{sc}
    \end{small}
    \vspace{-0.15in}
  \end{center}
    \caption{\textbf{Main Household Prediction Performance.}
 Comparison of ML and LLM baselines across travel-domain regression and mobility-domain classification. \textbf{Bold} denotes the best result within each dataset block. $\downarrow$/$\uparrow$ indicate lower/higher is better. $^\dag$Linear regression for NHTS/Puget Sound and logistic regression for PSID.}
    \label{tab:main-results}
  \vskip -0.1in
\end{table*}

\noindent\textbf{Baselines.} We benchmark PEMAND against conventional ML models, including Gradient Boosting and Random Forest, using identical features and data splits. We also compare against four LLM baseline families: \textbf{Demographics-Only}, which predicts directly from socioeconomic features; \textbf{Persona}, which applies the HA-CoPB prompt to the full household in one inference step; \textbf{RAG}, which augments Persona with retrieved similar households; and \textbf{Self-Consistency}, which aggregates over $K=10$ sampled predictions. Detailed prompts are provided in Appendix~\ref{app:prompts}.

\noindent \textbf{Datasets}
We evaluate PEMAND on three datasets across two domains: NHTS 2017 \cite{fhwa2017nhts} and Puget Sound 2023 \cite{psrc2023survey} for trip planning, and PSID 2023 \cite{psid2023dataset} for residential mobility. We use consistent preprocessing, variable selection, and train–test splits across datasets, with details in Appendix~\ref{app:datasets}.

\noindent \textbf{Evaluation Methods.} Because PEMAND covers both regression and classification tasks, we use domain-specific metrics. For trip planning on NHTS and Puget Sound, we report RMSE, MAE, sMAPE, and $\pm 2$ trip accuracy. For PSID relocation prediction, we report accuracy, F1, precision, and recall. Metric definitions are provided in Appendix~\ref{app: eva}.

We evaluate the predictive performance of PEMAND across three datasets and compare it with both traditional ML and LLM-based baselines. The results are reported in Table~\ref{tab:main-results}. Across both domains, the Persona baseline consistently outperforms the Demographics-only baseline. Persona reduces MAE from 4.06 to 3.75 on NHTS and from 3.31 to 2.89 on Puget Sound, while improving F1 from 0.40 to 0.44 on PSID. This suggests that HA-CoPB-based persona modeling helps capture behavioral antecedents more effectively than raw demographic features alone. RAG and Self-Consistency do not consistently outperform the single-pass Persona baseline, suggesting that generic retrieval or sampling strategies add limited value for novel household configurations. PEMAND achieves the best performance across all datasets, outperforming the strongest traditional baselines. It reduces MAE from 3.07 to 2.38 on NHTS and from 2.75 to 1.99 on Puget Sound, and improves F1 from 0.55 to 0.73 on PSID.

\subsection{Perception Validation of Synthesized Personas}
To assess persona coherence, we conduct a cross-domain \textit{Perception Validation} experiment using held-out Likert-scale probes. We evaluate \textit{Health Status}, \textit{Price Sensitivity}, and \textit{Walkability} in the travel domain, and \textit{Food Affordability}, \textit{Life Satisfaction}, and \textit{K6 Psychological Distress} in the mobility domain. This tests whether HA-CoPB personas infer psychologically plausible perceptions from demographics rather than simply reproducing input facts.

\begin{figure}[t]
    \centering
    \includegraphics[width=0.48\textwidth]{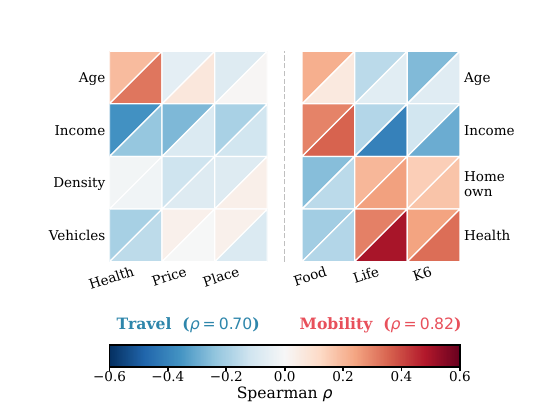}
    \caption{\textbf{Structural Correlation Alignment.} Diagonal-split heatmaps for demographic--opinion Spearman correlations in Travel (NHTS) and Mobility (PSID); upper-left triangle = human ground truth, lower-right = persona.}
    \vspace{-1em}
    \label{fig:perception_validation}
\end{figure}

As shown in Figure~\ref{fig:perception_validation}, HA-CoPB personas preserve latent demographic–opinion correlations in both domains, with Structural Correlation Alignment of $\rho=0.70$ for Travel and $\rho=0.82$ for Mobility. The figure compares human and persona Spearman correlations across key demographic variables, while individual calibration remains strong, with aggregate $\pm 1$ accuracies of 74.2\% on NHTS and 90.3\% on PSID. Full probe definitions, tabular results, and metric formulations are provided in Appendix~\ref{app:validation}.

\subsection{Ablation Study}
To assess each component, we conduct an ablation study on the Puget Sound 2023 dataset with five configurations: initial proposal without persona-alignment control, proposal with persona-based SFT, single agent with moderator and ReST, SFT-based household negotiation without moderator or ReST, negotiation with moderator control, and the full PEMAND model with ReST-based trajectory selection. 

The ablation results demonstrate the contribution of each component. Without SFT, the LLM-based agents exhibit substantial persona-alignment errors, including role hallucination, violations of household constraints, and invalid trip proposals. Introducing household negotiation improves over directly aggregating individual proposals, indicating that intra-household deliberation provides useful corrective signals. However, the gain remains limited without alignment control, since hallucinated or invalid proposals can propagate through the discussion and bias the final decision. Adding intra-conversation control substantially improves performance by filtering these failure modes and stabilizing the negotiation process. Finally, incorporating ReST further improves the full model by leveraging ground-truth supervision from the training set to select higher-quality discussion trajectories for fine-tuning.

\begin{table}[H]
  \begin{center}
    \scriptsize
    \setlength{\tabcolsep}{3pt}
    \begin{tabular}{l cccc}
      \toprule
      \textbf{Model}
      & MAE$\downarrow$ & RMSE$\downarrow$ & sMAPE$\downarrow$ & $\pm 2$ Acc$\uparrow$ \\
      \midrule
      Init. Proposal w/o SFT
      & 3.70 & 5.32 & 63.44 & 0.51 \\
      Init. Proposal w/ SFT
      & 3.10 & 4.47 & 57.90 & 0.56 \\
      Refinement w/o RL \& Moderator
      & 3.05 & 4.32 & 56.02 & 0.58 \\
      Refinement w/o RL
      & 2.59 & 3.91 & 53.80 & 0.64 \\
      \textbf{Consensus Refinement}
      & \textbf{1.99} & \textbf{3.25} & \textbf{47.92} & \textbf{0.78} \\
      \bottomrule
    \end{tabular}
      \caption{\textbf{Ablation Study on Consensus Refinement.}
  \textbf{Bold} indicates best performance.
  $\downarrow$ indicates lower is better, $\uparrow$ indicates higher is better.}
  \end{center}
  \label{tab:ablation-consensus}
  \vskip -0.1in
\end{table}

\section{Discussion - Why It Works}
The model’s ability to produce realistic and accurate household-level decisions stems from the careful design of each component in the proposed framework.

A key innovation is theory-guided, household-aware persona generation. Instead of using unconstrained textual personas, PEMAND grounds agents in TPB constructs and shared household context, enabling decision-relevant reasoning while preserving role and resource consistency. This turns persona generation into a structured behavioral abstraction, reducing hallucination and improving the plausibility of household decisions.

PEMAND decomposes household prediction into parallel individual proposal generation and consensus refinement. This structure reflects that household decisions emerge through interaction among members with different roles, constraints, and needs, rather than from simple aggregation. By separating initial preference elicitation from negotiation, the framework produces decisions that are both individually grounded and collectively consistent.

The framework further improves behavioral fidelity through intra- and out-of-conversation alignment control. By combining human-calibrated LLM judging, proposal-pair supervision, and ground-truth-guided trajectory selection, PEMAND reduces invalid or persona-inconsistent dialogue and refines agent policies toward more realistic household-level decisions.

\section{Conclusion}
We present PEMAND, a theory-guided multi-agent framework for modeling household-level decision-making with LLMs. By integrating household-aware persona generation, structured individual proposal elicitation, consensus-based refinement, and multi-level alignment control, PEMAND captures both individual heterogeneity and intra-household negotiation. Experiments across national and regional datasets show that PEMAND consistently improves predictive accuracy over traditional machine learning baselines and standard LLM-based approaches. These results suggest that structured, behaviorally grounded multi-agent LLM systems offer a promising direction for modeling collective human decisions in socially situated domains.

\section*{Limitations}

This work has several limitations. First, although PEMAND improves household-level prediction for both travel demand and relocation decisions, the negotiation process is mainly validated through outcome accuracy and persona-consistency checks, leaving richer behavioral dimensions such as trip timing, mode choice, bargaining dynamics, and relocation motivations for future study. Second, while PEMAND is designed around offline open-weight models and parallelizable inference, extending multi-agent negotiation to city-, regional-, or national-level simulation would benefit from further engineering optimizations, such as adaptive stopping, selective negotiation for high-uncertainty households, and batched deployment. Finally, PEMAND should be evaluated across more diverse household structures and extended to other collective decision-making domains, such as household economic decisions, fertility, labor supply, and residential choice.



\bibliography{custom}

\appendix

\section{Experimental Setup \& Baselines}

\subsection{Baseline Models}
\label{app:baseline}
To provide a comprehensive benchmark for the proposed agent-based approach, we established a set of conventional statistical and machine learning baseline models. These baselines were selected because they are widely used and cover a range of modeling assumptions and complexity, from interpretable linear models to flexible nonlinear learners.

The statistical baselines include linear regression, Poisson regression, and negative binomial regression, which are commonly applied to count-based travel outcomes. Poisson regression serves as a standard framework for modeling trip counts, while negative binomial regression is included to account for potential over-dispersion in household trip frequency. Furthermore, for the Mobility Domain (PSID), logistic regression was used to baseline the binary decision.

In addition, we evaluated several machine learning baselines, including random forest regression, gradient boosting regression, and a multi-layer perceptron (MLP) regressor/classifier, which can capture nonlinear relationships and higher-order interactions among socio-demographic, built environment, and contextual features without requiring strong distributional assumptions.

\subsection{Implementation and Computational Details}
\label{app:implementation}
\subsubsection{Supervised Fine-Tuning (SFT) Hyperparameters}
For the supervised fine-tuning phase, we employed QLoRA to efficiently adapt Llama-3.1-8B-Instruct while keeping the backbone frozen and quantized to 4-bit (NF4) with double quantization and bfloat16 compute dtype. LoRA adapters were attached to the attention and MLP projections (q\_proj, k\_proj, v\_proj, o\_proj, gate\_proj, up\_proj, down\_proj) using a rank of 16, an alpha of 32, and a dropout rate of 0.05. We utilized the paged 8-bit AdamW optimizer with a learning rate of $2 \times 10^{-4}$, a cosine learning-rate schedule with a 3\% warmup ratio, and an effective batch size of 8 (per-device batch size of 2 with 4 gradient-accumulation
steps). Training was conducted for 3 epochs with a maximum sequence length of 1024 tokens, gradient checkpointing enabled, bfloat16 mixed precision, and a fixed random seed of 0. Checkpoints were saved at the end of each epoch, retaining the two most recent. Fine-tuning was executed on a single GPU with approximately 24 GB of memory.

\subsection{Parameter Setting}
We set the maximum number of discussion rounds to $R_{\max}=5$ and the per-turn revision clamp to $\rho_{\text{turn}}=0.5$. The downward and upward bounding parameters are set to $\rho_{\text{down}}=1.0$ and $\rho_{\text{up}}=0.4$, respectively. For reward weighting, we set the ground-truth term to $\omega_{GT}=2.0$ and the rubric-based term to $\omega_{R}=0.5$.

\subsubsection{Inference Cost and Latency}
The multi-agent framework requires several sequential steps per household, including persona proposal generation, moderator validation, and iterative discussion. Across the evaluation datasets, processing a single household required, on average, 12.72, 16.31, and 17.88 model calls for NHTS 2017, Puget Sound 2023, and PSID, respectively. The corresponding average input-token usage was 8,854.60, 13,324.90, and 19938.4, while the average output-token usage was 2,029.24, 3,561.37, and 2447.4. The average end-to-end inference runtime per household was 19.65, 22.54, and 40.02 seconds, respectively. While this is more expensive than a one-shot prediction baseline, the cost reflects the additional computational complexity required to model coordinated, constraint-aware household decision-making rather than relying on a monolithic generation.

\section{Dataset and Preprocessing Details}
\label{app:datasets}

\subsection{Dataset Descriptions}
\subsubsection{NHTS 2017}
We use the household and person files from the National Household Travel Survey (NHTS) 2017 \cite{fhwa2017nhts}, a nationally representative travel survey conducted by the U.S. Federal Highway Administration. The person file contains approximately 264,235 records corresponding to individual household members, while the household file includes approximately 129,697 household-level records.

Explanatory variables are selected to capture factors known to influence household trip frequency, including socio-demographics (e.g., household composition and age structure), household attributes (e.g., income, household size, and vehicle ownership), built environment characteristics (e.g., land-use mix and density measures), and economic context variables (e.g., travel-day gas prices). Built environment and economic context variables are available only in the NHTS dataset.

Categorical variables are transformed using one-hot encoding. Since all baseline models operate at the household level, person-level attributes are aggregated into household-level features using within-household proportions (e.g., percentage of workers or students) or averages (e.g., physical activity level).

For model evaluation, we use a 90/10 train--test split, following common practice in large-scale travel behavior modeling.

\subsubsection{Puget Sound 2023}
To assess generalizability, we include a regional case study using the Puget Sound household travel survey (2023) \cite{psrc2023survey}. The dataset contains approximately 7,562 person records and 3,871 household records. We apply the same variable selection, feature engineering, and preprocessing pipeline as used for the NHTS dataset, excluding built environment and economic context variables that are not available in this survey.

Models are evaluated using an 80/20 train--test split, reflecting the smaller sample size of the regional dataset.

\subsubsection{Panel Study of Income Dynamics (PSID) 2023}
We use the 2023 family file and the 1968 - 2023 individual files from the Panel Study of Income Dynamics (PSID), a longitudinal household panel survey that has followed U.S. families and their descendants since 1968. The 2023 family file contains 9,152 household-level records, while the individual file contains 85,536 person records. The prediction task is binary classification of target variable:  \textbf{residential mobility}, which means that a household changed their residence between 2021 and 2023 PSID waves. The processed household level dataset contains 9152 households, with approximately 30\% of the households labeled as they moved recently. The processed PSID features are prepared by removing identifiers, redundant locations, encoding categorical variables, scaling continuous variables, and retaining ordinal and percentage based household measures in their original form.

\subsection{Data Preprocessing and Imputation}

\subsubsection{Missing-Data Handling}
Because missing and ``skip-coded'' values are common in survey data, we applied multiple strategies to handle incomplete records across all datasets:
\begin{enumerate}
    \item \textbf{Household-level removal:} If any household member contained invalid or missing values for critical variables, the entire household was excluded to maintain consistency in household-level aggregation.
    \item \textbf{Rule-based recoding:} Selected missing or negative-coded responses were recoded using domain-informed rules. For example, negative codes for rideshare usage were treated as no rideshare use.
    \item \textbf{Median/mode imputation:} For remaining variables with missing values, numeric features were imputed using the median and categorical features were imputed using the mode to reduce data loss while preserving overall distributions.
\end{enumerate}

\subsubsection{Feature Engineering and Scaling}
For the classical machine learning baselines, datasets were preprocessed using the following steps:
\begin{itemize}
    \item \textbf{Categorical Encoding:} Categorical attributes (excluding those already ordinal encoded) were transformed using one-hot encoding.
    \item \textbf{Aggregation (Travel Domain):} Since all baseline models operate at the household level, person-level attributes in the NHTS and Puget Sound datasets were aggregated into household-level features using within-household proportions (e.g., percentage of workers or students) or averages (e.g., physical activity level).
    \item \textbf{Scaling (Mobility Domain):} For the PSID dataset, identifiers and redundant locations were removed. Percentage features were retained in their original form. The remaining numerical features were first power-transformed using the Yeo-Johnson method before standardization to handle skewness.
\end{itemize}

\subsection{Variable Definitions}
Table~\ref{tab:variables} lists the variables used in the travel domain experiments. Variables marked with ``*'' are included in both the NHTS and Puget Sound datasets. Table~\ref{tab:psid_variables} lists the variables utilized for the processed PSID mobility dataset.

\subsection{Coding Schemes}
The coding conventions used for binary, frequency-based, and income bracket variables are summarized in Table~\ref{tab:coding}.

\begin{table*}[h!]
\centering
\small
\begin{tabular}{lll}
\hline
\textbf{Variable Type} & \textbf{Code / Value} & \textbf{Meaning} \\
\hline
Binary & 0 & No \\
Binary & 1 & Yes \\
\hline
Frequency & 01 & Never \\
Frequency & 02 & A few times a year \\
Frequency & 03 & A few times a month \\
Frequency & 04 & A few times a week \\
Frequency & 05 & Daily \\
\hline
Income Bracket & 1 & Less than \$25k \\
Income Bracket & 2 & \$25k -- \$50k \\
Income Bracket & 3 & \$50k -- \$75k \\
Income Bracket & 4 & \$75k -- \$100k \\
Income Bracket & 5 & Higher than \$100k \\
\hline
\end{tabular}
\caption{Coding schemes for binary, frequency, and income bracket variables.}
\label{tab:coding}
\end{table*}

\subsection{Evaluation Metrics Details}
\label{app: eva}
We compare predicted household trip counts with observed values using multiple complementary evaluation metrics, including Root Mean Squared Error (RMSE), Mean Absolute Error (MAE), symmetric Mean Absolute Percentage Error (sMAPE), and a tolerance-based accuracy measure.
Together, these metrics capture overall goodness-of-fit, absolute and relative error magnitudes, and the practical usefulness of predictions for household-level travel decision-making and transportation planning contexts.

Furthermore, we used accuracy, precision, recall, and F1 score for binary classification evaluations.

\paragraph{Root Mean Squared Error (RMSE)}
RMSE measures the square root of the average squared difference between predicted and actual values:
\begin{equation}
\text{RMSE} = \sqrt{\frac{1}{n}\sum_{i}(y_i - \hat{y}_i)^2}.
\end{equation}
By squaring the errors, RMSE penalizes large deviations more heavily and is therefore sensitive to extreme prediction errors.

\paragraph{Mean Absolute Error (MAE)}
MAE computes the average absolute difference between predicted and actual values:
\begin{equation}
\text{MAE} = \frac{1}{n}\sum_{i}|y_i - \hat{y}_i|.
\end{equation}
Compared to RMSE, MAE is less influenced by outliers and provides an interpretable measure of the typical prediction error in the original unit of trips.

\paragraph{Symmetric Mean Absolute Percentage Error (sMAPE)}
sMAPE is a scale-independent metric that evaluates relative prediction error:
\begin{equation}
\text{sMAPE} = \frac{1}{n}\sum_{i}
\frac{2|y_i - \hat{y}_i|}{|y_i| + |\hat{y}_i|}.
\end{equation}
Household trip counts exhibit a high frequency of zero-valued observations, for which conventional percentage-based error metrics such as MAPE become undefined or numerically unstable.
By symmetrically normalizing prediction errors using both predicted and actual values, sMAPE provides a more stable evaluation of relative error in zero-inflated travel demand data.

\paragraph{Accuracy within a Tolerance ($\pm 2$ Trips)}
To assess practical prediction performance, we report a tolerance-based accuracy metric defined as the proportion of households whose predicted trip counts fall within $\pm 2$ trips of the actual values.
Formally, a prediction for household $i$ is considered accurate if:
\begin{equation}
|y_i - \hat{y}_i| \leq 2,
\end{equation}
and the reported accuracy is computed as the fraction of households satisfying this condition across the evaluation set.
Household-level trip counts are obtained by aggregating discrete individual trip counts across household members, most of whom form households of two or more individuals.

As a result, small deviations at the individual level naturally translate into differences of one to two trips at the household level, making a $\pm 2$ tolerance a reasonable margin of error given the data aggregation process.

\paragraph{Accuracy}
Accuracy measures the proportion of correctly classified instances among all evaluated cases:
\begin{equation}
\text{Accuracy} = \frac{TP + TN}{TP + TN + FP + FN}.
\end{equation}
Accuracy provides an intuitive summary of overall classification performance but can be misleading under class imbalance, where a trivial majority-class predictor may still attain a high score.

\paragraph{Precision}
Precision quantifies the proportion of predicted positives that are truly positive:
\begin{equation}
\text{Precision} = \frac{TP}{TP + FP}.
\end{equation}
Precision is informative when false positives are costly, as it reflects how reliable a positive prediction is.

\paragraph{Recall}
Recall measures the proportion of actual positives that are correctly identified:
\begin{equation}
\text{Recall} = \frac{TP}{TP + FN}.
\end{equation}
Recall is particularly relevant when false negatives carry a high cost, as it captures how completely the model retrieves positive cases.

\paragraph{F1 Score}
The F1 score is the harmonic mean of precision and recall, balancing the two into a single summary statistic:
\begin{equation}
F_1 = \frac{2 \cdot \text{Precision} \cdot \text{Recall}}{\text{Precision} + \text{Recall}}.
\end{equation}
By combining precision and recall, the F1 score is particularly useful under class imbalance, where accuracy alone may overstate performance.

\section{Additional Main Results}
\label{app:additional_main}

\subsection{Robustness Across Different LLM Backbones}\label{app:robustness}
\begin{table*}[t]
  \centering
  \small
  \setlength{\tabcolsep}{5pt}
  \resizebox{\textwidth}{!}{
  \begin{tabular}{l rrrr rrrr rrrr}
    \toprule
    & \multicolumn{4}{c}{\textbf{Llama-3.1-8B-Instruct}} & \multicolumn{4}{c}{\textbf{Gemma-3-27b-it}} & \multicolumn{4}{c}{\textbf{GPT-5\_4-mini}}\\
    \cmidrule(lr){2-5} \cmidrule(lr){6-9} \cmidrule(lr){10-13}
    Method
    & MAE$\downarrow$ & RMSE$\downarrow$ & sMAPE$\downarrow$ & $\pm2$ Acc$\uparrow$
    & MAE$\downarrow$ & RMSE$\downarrow$ & sMAPE$\downarrow$ & $\pm2$ Acc$\uparrow$
    & MAE$\downarrow$ & RMSE$\downarrow$ & sMAPE$\downarrow$ & $\pm2$ Acc$\uparrow$\\
    \midrule
    Demographics-only
      & 3.31 & 5.10 & 61.75 & 0.54
      & 2.83 & 4.19 & 55.63 & 0.58 
      & 2.98 & 4.23 & 56.47 & 0.55\\
    Persona
      & 2.89 & 3.99 & 59.52 & 0.56
      & 3.33 & 5.37 & 62.99 & 0.55 
      & 3.09 & 4.25 & 56.13 & 0.53\\
    RAG
      & 5.01 & 6.60 & 76.49 & 0.35
      & 3.93 & 5.43 & 66.85 & 0.43 
      & 3.46 & 4.71 & 60.61 & 0.47\\
    Self-consistency
      & 2.89 & 4.04 & 58.30 & 0.57
      & 3.01 & 4.27 & 61.70 & 0.55
      & 2.91 & 4.07 & 54.34 & 0.57\\
    PEMAND w/o ReST
      & 2.59 & 3.91 & 53.80 & 0.64
      & 2.41 & 3.64 & 49.54 & 0.68
      & \textbf{2.32} & \textbf{3.78} & \textbf{47.87} & \textbf{0.63}\\
    PEMAND w ReST
      & \textbf{1.99} & \textbf{3.25} & \textbf{47.92} & \textbf{0.78}
      & \textbf{1.68} & \textbf{2.98} & \textbf{45.73} & \textbf{0.81} 
      & --- & --- & --- & ---\\
    \bottomrule
  \end{tabular}}
  \vspace{0.3em}
  \begin{minipage}{\textwidth}\footnotesize
  \textbf{Note:} We do not report \textbf{PEMAND w ReST} for GPT-5.4-mini because QLoRA adapter training was applied only to open-weight backbones; GPT results use API inference without weight updates.
  \end{minipage}
  \caption{\textbf{LLM Baselines on Puget Sound 2023 (test).}
    Comparison of \textbf{Llama-3.1-8B-Instruct}, \textbf{Gemma-3-27b-it}, and \textbf{GPT-5\_4-mini} under the same baseline families, with a \emph{homogeneous} backbone in each column (persona generation, member agents, moderator, and judge share the same model family).
    \textbf{Bold} is best among the four LLM baselines within each model block and metric column.
    \textbf{PEMAND w/o ReST} / \textbf{PEMAND w ReST}: multi-agent pipeline without or with QLoRA SFT on member models for that backbone column.
    $\downarrow$ / $\uparrow$ indicate lower / higher is better.}
\label{tab:puget-baselines}
\end{table*}

\begin{figure*}[t]
\centering
\includegraphics[width=0.92\textwidth]{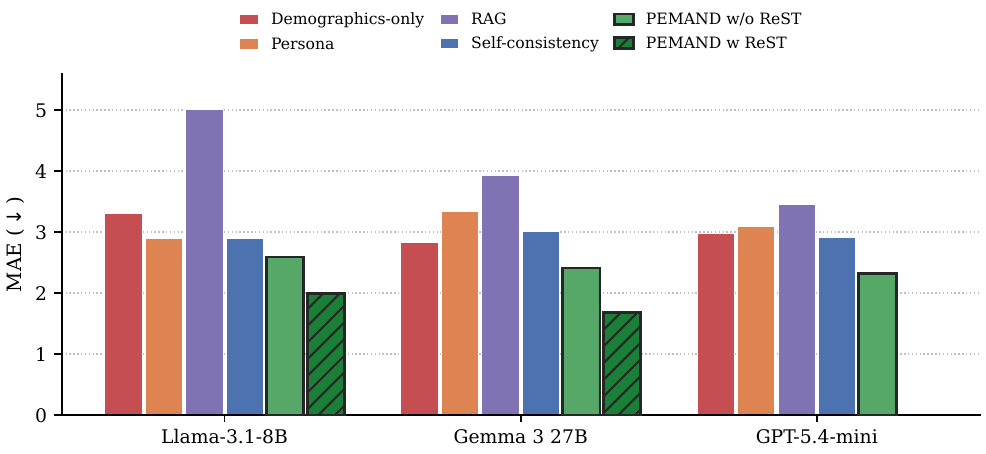}
\caption{\textbf{Robustness across homogeneous LLM backbones (Puget Sound test, MAE).}
  Grouped bars compare baseline families under Llama-3.1-8B-Instruct, Gemma-3-27b-it, and GPT-5.4-mini.
  PEMAND achieves the lowest MAE on every backbone; stronger models reduce error for most families, though persona conditioning is not uniformly beneficial.
  PEMAND w ReST is omitted for GPT-5.4-mini (QLoRA training applied only to open-weight backbones; Table~\ref{tab:puget-baselines}).}
\label{fig:robustness_backbones}
\end{figure*}

To evaluate the robustness of our framework across different model architectures, we conducted a side-by-side comparison deploying homogeneous pipelines end-to-end using Llama-3.1-8B-Instruct, Gemma-3-27b-it, and GPT-5.4-mini on the Puget Sound dataset (Table \ref{tab:puget-baselines}). The results demonstrate that PEMAND consistently outperforms all single-agent baseline variants regardless of the underlying LLM backbone. We observe several key insights:

First, performance across zero-shot single-agent baselines are highly model-dependent. While the Demographics-only baseline struggles on Llama (MAE 3.31), it performs well on Gemma (MAE 2.83, $\pm2$ accuracy 0.58), surpassing both Llama and GPT-5.4-mini. Second, single-agent persona conditioning is not uniformly beneficial. While injecting a behavioral persona improves Llama's prediction (reducing MAE to 2.89), it degrades performance for both Gemma (MAE 3.33) and GPT-5.4-mini (MAE 3.09) compared to their respective Demographics-only baselines. Retrieval-Augmented Generation (RAG) remains universally harmful across all three backbones, likely because retrieving historical households with conflicting trip counts induces severe covariate shift. By contrast, Self-consistency remains a strong single-agent strategy, particularly for GPT-5.4-mini (MAE 2.91) and Gemma (MAE 3.01), by reducing sampling variance.

Ultimately, the complete PEMAND multi-agent negotiation framework achieves the best overall performance across backbone configurations. The Gemma-3-27B model yields stronger overall results than Llama-3.1-8B, which is expected given its larger parameter scale and stronger generation capacity. Importantly, however, PEMAND remains highly competitive even with the lighter Llama backbone, indicating that the performance gains do not rely solely on model size. Instead, the results suggest that the structured negotiation protocol, alignment control, and trajectory refinement provide robust error reduction across different LLM backbones. This supports the generality of the proposed framework and shows that PEMAND can benefit from larger frontier-scale models while remaining effective under more lightweight and deployable open-weight settings.

\section{Prompt Engineering Interfaces}
\label{app:prompts}

We formalize our prompt designs as structured cognitive interfaces. Each template employs explicit role definition, context delimitation, and negative constraint protocols to ensure reproducibility and mitigate LLM hallucination.

\textbf{Domain Generalization Mapping:} To support multiple decision domains, our prompt architectures use unified structural placeholders that are dynamically instantiated based on the target dataset. Table \ref{tab:prompt_domain_mapping} illustrates the mappings across the two primary case studies.

\begin{table*}[t] 
\centering
\small
\begin{tabular}{@{}lll@{}}
\toprule
\textbf{Placeholder} & \textbf{Travel Domain (NHTS/Puget)} & \textbf{Mobility Domain (PSID)} \\
\midrule
\texttt{<target decision>} & Total daily trip frequency & Relocated between 2021--2023 \\
\texttt{<shared constraints>} & Vehicle deficit/bottlenecks & Homeownership lock-in \\
\texttt{<agent role>} & Travel Diary Participant & Head of Household/Spouse \\
\texttt{<output format>} & Integer (e.g., 4) & Boolean (True/False) \\
\bottomrule
\end{tabular}
\caption{Dynamic prompt instantiation across domains.}
\label{tab:prompt_domain_mapping}
\end{table*}

\subsection{Phase 1: Faithfulness-Constrained Narrative Synthesis}
The following prompts drive the transformation of tabular encoded data into natural language personas. We employ a Clinical Neutrality Protocol to strip RLHF-induced positivity bias. To prevent the persona from embedding target decision outcomes, we impose task-specific vocabulary restrictions. For example, in the residential mobility task, we explicitly ban evaluative adjectives (e.g., "stable", "rooted", "transient") and predictive language (e.g., "likely to move", "intends to relocate") to avoid outcome-leakage or stereotyping. We also enforce relationship-aware constraints to ensure syntactical consistency with the individual's role relative to the household head.

\begin{tcolorbox}[colback=gray!5!white,colframe=gray!75!black,title=Box 1: Persona Generation System Definition]
\small\ttfamily
\textbf{[ROLE DEFINITION]}\\
You are a factual data-to-text transcription engine. Your objective is to convert atomic variable states into a coherent first-person narrative without semantic alteration or emotional embellishment.

\textbf{[STRICT CONSTRAINTS]}\\
1. \textbf{Fidelity}: Use every provided fact exactly as given. Do not hallucinate traits not present in the input.\\
2. \textbf{Sentiment Neutralization}: The following subjective descriptors are \textbf{BANNED}: \textit{proud, blessed, fortunate, thrilled, excited, grateful}.\\
3. \textbf{Role Rigor}: Do not infer authority phrases like "Head of Household" or "responsible for" unless explicitly encoded in the relationship variable.\\
4. \textbf{Vocabulary Restrictions}: \textit{stable, rooted, transient, likely to move, intends to relocate} are \textbf{BANNED}.\\
5. \textbf{Scope}: NEVER mention the target variable. Output ONLY the persona text.
\end{tcolorbox}

\begin{tcolorbox}[colback=gray!5!white,colframe=gray!75!black,title=Box 2: Persona Generation Input Interface]
\small\ttfamily
\textbf{[INPUT DATA: ATOMIC FACTS]}\\
The following list represents the ground-truth state of the agent:\\
\{facts\}

\textbf{[GENERATION INSTRUCTIONS]}\\
Synthesize a first-person narrative adhering to the following logic:\\
- \textbf{Identity}: Begin with "I am..."\\
- \textbf{Completeness}: Integrate 100\% of the atomic facts listed above.\\
- \textbf{Tone}: Maintain a clinical, objective tone. Avoid filler words.\\
- \textbf{Consistency}: If facts imply conflicting states (e.g., high income but zero car), state them clearly without attempting to "fix" the contradiction.

\textbf{[OUTPUT]}\\
Persona:
\end{tcolorbox}

\subsection{Phase 2: Individual Reasoning \& Alignment via Refinement (AvR)}
These templates drive the core HA-CoPB reasoning engine and the dual-agent alignment layer. The Actor generates the TPB evaluation, and the Reviewer calibrates it against the statistical prior.

\begin{tcolorbox}[colback=gray!5!white,colframe=gray!75!black,title=Box 3: HA-CoPB Actor Reasoning Template]
\small\ttfamily
\textbf{[SYSTEM CONTEXT]}\\
You are a behavioral reasoning engine. Your goal is to predict the \texttt{<target decision>} for a specific household member, but you must account for the shared constraints and obligations of the entire household.

\textbf{[HOUSEHOLD STATE]}\\
\textbf{Context}: Size: \{hh\_size\} | Vehicles: \{veh\_count\} | Income: \{income\}\\
\textbf{Shared Constraints}: \texttt{<shared constraints>}\\
\textbf{Behavioral Anchor}: Similar individuals typically exhibit \textbf{\{anchor\}} behavior.

\textbf{[PERSONA PROFILE]}\\
\textbf{Persona}: \{persona\}\\
\textbf{Behavioral Tendency}: \{marker\_statement\} (Implication: \{marker\_implication\})

\textbf{[HA-CoPB REASONING TASK]}\\
Perform a Theory of Planned Behavior (TPB) analysis that entangles individual desires with household reality:

\textbf{1. Attitude (Individual Utility)}: \\
   - How does the "Behavioral Tendency" shift their intrinsic demand?

\textbf{2. Subjective Norms (Household Obligations)}: \\
   - What role-based obligations exist?

\textbf{3. Perceived Behavioral Control (Resource Competition)}: \\
   - Do financial, spatial, or shared resource constraints limit their autonomy?

\textbf{[OUTPUT]}\\
\textbf{Rationale}: $<$Step-by-step TPB analysis resolving the conflicts above$>$
\end{tcolorbox}

\begin{tcolorbox}[colback=gray!5!white,colframe=gray!75!black,title=Box 4: AvR Moderator Calibration Template]
\small\ttfamily
\textbf{[SYSTEM CONTEXT]}\\
You are the Lead Demographic Sociologist and Calibration Moderator. Your task is to review the Theory of Planned Behavior reasoning generated by the primary behavioral analyst and issue the final calibrated individual intention.

\textbf{[EMPIRICAL BASELINE (ANCHOR)]}\\
Households with demographics similar to this one have an empirical baseline rate of \textbf{\{anchor\}}. 

\textbf{[CALIBRATION RULES]}\\
1. Treat this empirical anchor as a qualitative prior expectation, NOT a strict mathematical threshold.\\
2. Read the primary analyst's reasoning. If the analyst identifies overwhelming desires that overpower the macro-trend, you MUST deviate from a low baseline.\\
3. If the analyst identifies absolute structural vetoes (like \texttt{<shared constraints>}), you MUST enforce the constraint, regardless of their positive attitude.

\textbf{[OUTPUT]}\\
Provide a brief justification evaluating if the analyst's identified constraints or facilitators warrant the action, then output:\\
\textbf{Final Answer}: \texttt{<output format>}
\end{tcolorbox}

\subsection{Phase 3: Multi-Agent Parallel Proposal}
These templates illustrate how member agents express their initial intentions and needs during the proposal phase.
\label{sub: parallel}

\begin{tcolorbox}[breakable, colback=gray!5!white,colframe=gray!75!black,title=Box 5: Parallel Proposal Prompt Template]
  \small\ttfamily
  \textbf{[SYSTEM CONTEXT]}\\
  Role-play as one household member. Stay in character using the given sociodemographic profile and TPB perceptions. Respect household constraints and do not fabricate members or resources.

  \textbf{[AGENT PERSONA]}\\
  \texttt{<sociodemographic profile>}

  \textbf{[TPB PERCEPTIONS]}\\
  \texttt{<TPB perceptions>}

  \textbf{[TASK]}\\
  Independently propose your initial \texttt{<decision>}. Explain the reasoning in 2--4 persona-grounded sentences, then output the decision on the last line.

  \textbf{[OUTPUT FORMAT]}\\
  Respond \textbf{STRICTLY} as:\\
  \texttt{REASON: <2--4 sentence justification>}\\
  \texttt{FINAL\_<DECISION>: <value>}
\end{tcolorbox}

\subsection{Phase 4: Multi-Agent Consensus Refinement}
These templates illustrate how member agents interact, revise their proposals in response to other members’ needs and opinions, and move toward a shared household decision.
\label{sub: refinement}
\begin{tcolorbox}[breakable, colback=gray!5!white,colframe=gray!75!black,title=Box 6: Consensus Refinement]
  \small\ttfamily
  \textbf{[SYSTEM CONTEXT]}\\
  You are one household member discussing a shared household-level \texttt{<decision>}. Stay in character and revise your view when another member provides valid evidence.

  \textbf{[YOUR IDENTITY]}\\
  \texttt{<persona + TPB perceptions + initial proposal>}

  \textbf{[OTHER MEMBERS]}\\
  \texttt{<other members' personas + initial proposals>}

  \textbf{[HOUSEHOLD CONSTRAINTS]}\\
  \texttt{<household size, income, vehicles, drivers, and non-speaking dependents>}

  \textbf{[STARTING POINT]}\\
  Initial values: \texttt{<list>}\\
  Aggregated initial \texttt{<decision>}: \texttt{<$\hat{y}_0$>}

  \textbf{[CURRENT VALUES]}\\
  \texttt{<latest PREFERRED\_<DECISION> from each speaker>}

  \textbf{[CONVERSATION HISTORY]}\\
  \texttt{<$h_t$ — truncated dialogue history>}

  \textbf{[TASK]}\\
  Reassess all members' evidence and either propose a revised household \texttt{<decision>} or use \texttt{SKIP} after your first turn. Acknowledge a specific point from another member, justify any change from your previous value, and respect household constraints.

  \textbf{[OUTPUT FORMAT]}\\
  Respond \textbf{STRICTLY} as:\\
  \texttt{ACKNOWLEDGE: <specific point from another member or initial value>}\\
  \texttt{SUPPORT: <evidence-grounded justification>}\\
  \texttt{PREFERRED\_<DECISION>: <value>}\\
  \texttt{DELTA\_FROM\_LAST: <brief reason for change or no change>}\\
  Or after the first turn: \texttt{SKIP}.
\end{tcolorbox}

\subsection{Phase 5: Discussion Moderator}
These templates show how the moderator evaluates each member agent’s utterance.
\label{app: moderator}
\begin{tcolorbox}[breakable, colback=gray!5!white,colframe=gray!75!black,title=Box 8: Moderator Validation Prompt Template]
\small\ttfamily
\textbf{[SYSTEM CONTEXT]}\\
You are the moderator agent supervising a household decision-making discussion. Your role is to evaluate whether a candidate utterance satisfies the provided annotation guideline. You do \textbf{not} participate as a household member and do \textbf{not} revise the utterance.

\textbf{[ANNOTATION GUIDELINE]}\\
\texttt{<annotation guideline>}

\textbf{[HOUSEHOLD CONTEXT]}\\
\texttt{<household context>}

\textbf{[SPEAKER PROFILE]}\\
\texttt{<agent persona profile>}

\textbf{[HOUSEHOLD PROPOSAL]}\\
Current household decision estimate: <$\hat{y}_t$>

\textbf{[CONVERSATION HISTORY]}\\
<$h_t$>

\textbf{[CANDIDATE UTTERANCE]}\\
<$u_{candidate}$>

\textbf{[TASK: MODERATOR VALIDATION]}\\
Evaluate the candidate utterance according to the provided annotation guideline. Assign a score from 1 to 3 for each rubric dimension, where 3 indicates full compliance with the guideline. The candidate utterance passes validation only if it receives a score of 3 on every dimension. If any dimension receives a score below 3, the utterance must be rejected and regenerated. Rejected utterances are discarded and \textbf{not} added to the conversation history.

\textbf{[OUTPUT FORMAT]}\\
Respond with:\\
\texttt{SCORES: <dimension\_1>=<1--3>; <dimension\_2>=<1--3>; ...}\\
\texttt{DECISION: PASS} \quad or \quad \texttt{DECISION: REJECT}\\
\texttt{FEEDBACK: <one short reason if rejected; otherwise "All dimensions receive full scores.">}
\end{tcolorbox}

\subsection{Phase 6: Comparison Baselines}

\subsubsection{Baseline: Demographics Only}
\begin{tcolorbox}[breakable, colback=gray!5!white,colframe=gray!75!black,title=Box 8: Demographics-Only Baseline Interface]
\small\ttfamily
\textbf{[SYSTEM ROLE]}\\
You function as a direct feature-to-label predictor.

\textbf{[INPUT FEATURES]}\\
\textbf{Demographic Profile}:\\
\{demographics\}

\textbf{[PREDICTION TASK]}\\
Based strictly on the features provided above, predict the \texttt{<target decision>} for this agent.

\textbf{[OUTPUT]}\\
Provide a single-sentence rationale followed by the prediction.\\
\textbf{Rationale}: $<$Brief justification$>$\\
\textbf{Final Answer}: \texttt{<output format>}
\end{tcolorbox}

\subsubsection{Baseline: Persona}
\begin{tcolorbox}[breakable, colback=gray!5!white,colframe=gray!75!black,title=Box 9: Persona Baseline]
\small\ttfamily
\textbf{[SYSTEM CONTEXT]}\\
You are an expert Behavioral Psychologist predicting \textbf{total household behavior}.

\textbf{[MEMBER PROFILES]}\\
The following narratives describe the household members:\\
\{member\_narratives\}

\textbf{[PREDICTION TASK]}\\
Predict the \texttt{<target decision>} using Theory of Planned Behavior. You must mentally simulate the negotiation of shared constraints:

\textbf{1. Aggregated Needs (Attitude)}: \\
   - Sum the mandatory needs for all members.

\textbf{2. Shared Constraints (PBC)}: \\
   - \textit{Bottleneck Check}: Evaluate shared resource limitations.

\textbf{3. Synthesis}: \\
   - Start with the Anchor (\{anchor\}).\\
   - Adjust based on the net balance of Needs vs. Constraints.\\

\textbf{[OUTPUT]}\\
Respond STRICTLY in this format:\\
\textbf{Needs Analysis}: $<$Who needs the action?$>$\\
\textbf{Constraint Logic}: $<$How do limits reduce the probability?$>$\\
\textbf{Final Answer}: \texttt{<output format>}
\end{tcolorbox}

\subsubsection{Baseline: Self-Consistency}
The Self-Consistency baseline does not employ a distinct prompt structure. Instead, it utilizes the \textit{Persona} template (Box 9), but executes the inference step $K=10$ times at a non-zero temperature ($T=0.7$). The final prediction is generated by taking the majority vote (for classification tasks) or the median (for regression tasks) of the $K$ sampled outputs. This baseline tests whether stochastic exploration of reasoning paths improves zero-shot predictive alignment without requiring explicit multi-agent iteration.

\subsubsection{Baseline: Retrieval-Augmented Generation (RAG)}
The RAG baseline augments the Household CoPB prompt by injecting $N=3$ historically retrieved household exemplars (nearest neighbors in the training set based on demographic feature similarity) to provide local context.

\begin{tcolorbox}[colback=gray!5!white,colframe=gray!75!black,title=Box 10: RAG Baseline Template]
\small\ttfamily
\textbf{[SYSTEM CONTEXT]}\\
You are an expert Behavioral Psychologist predicting \textbf{total household behavior}.

\textbf{[HISTORICAL EXEMPLARS (RETRIEVED CONTEXT)]}\\
The following similar households from the historical database made the following decisions:\\
- \textit{Exemplar 1}: \{retrieved\_hh\_1\_features\} $\rightarrow$ \textbf{Outcome: \{retrieved\_outcome\_1\}}\\
- \textit{Exemplar 2}: \{retrieved\_hh\_2\_features\} $\rightarrow$ \textbf{Outcome: \{retrieved\_outcome\_2\}}\\
- \textit{Exemplar 3}: \{retrieved\_hh\_3\_features\} $\rightarrow$ \textbf{Outcome: \{retrieved\_outcome\_3\}}\\

\textbf{[CURRENT HOUSEHOLD: MEMBER PROFILES]}\\
The following narratives describe the current household members:\\
\{member\_narratives\}

\textbf{[PREDICTION TASK]}\\
Predict the \texttt{<target decision>} using Theory of Planned Behavior. Leverage the historical exemplars above as context for how similar households resolve their constraints.

\textbf{[OUTPUT]}\\
Respond STRICTLY in this format:\\
\textbf{Needs Analysis}: $<$Who needs the action?$>$\\
\textbf{Constraint Logic}: $<$How do limits reduce the probability?$>$\\
\textbf{Final Answer}: \texttt{<output format>}
\end{tcolorbox}

\section{Behavioral Theory Implementation}
\label{app:theory}

This section details the formal extraction and reasoning rules used to map raw survey data into theory-grounded cognitive representations for both the Travel Behavior (NHTS/Puget Sound) and Residential Mobility (PSID) domains.

\subsection{HA-CoPB Framework Overview}
\label{app:tpb_diagram}
Figure~\ref{fig:tpb_diagram} illustrates how our Household-Aware CoPB (HA-CoPB)
framework adapts the classical Theory of Planned Behavior (TPB) \cite{ajzen1991theory}
to the household decision-making setting. While the standard TPB models behavioral
intention as a function of Attitude, Subjective Norms, and Perceived Behavioral
Control at the individual level, our HA-CoPB reinterprets these constructs to
capture household-level dynamics:
$\mathcal{PBC}$ captures shared resource bottlenecks, $\mathcal{SN}$ captures concrete
intra-household role obligations, and $\mathcal{A}$ captures individual intrinsic preferences.

\begin{figure*}[htbp]
\centering
\begin{tikzpicture}[
    node distance=1.8cm and 4.2cm,
    box/.style={rectangle, draw, rounded corners, minimum width=2.8cm,
                minimum height=0.9cm, align=center, font=\small},
    arrow/.style={-{Stealth[length=2.5mm]}, thick},
    label/.style={font=\scriptsize\itshape, text=gray}
]
\node[box, fill=blue!10] (att) {Attitude\\($\mathcal{A}$)};
\node[box, fill=green!10, below=of att] (sn) {Subjective\\Norms ($\mathcal{SN}$)};
\node[box, fill=red!10, below=of sn] (pbc) {Perceived Behavioral\\Control ($\mathcal{PBC}$)};
\node[box, fill=yellow!15, right=of sn] (intent) {Behavioral\\Intention};
\node[box, fill=orange!15, right=of intent] (behav) {Behavior};

\draw[arrow] (att) -- (intent);
\draw[arrow] (sn) -- (intent);
\draw[arrow] (pbc) -- (intent);
\draw[arrow] (pbc) to[bend right=25] (behav);
\draw[arrow] (intent) -- (behav);

\node[above=0.3cm of att, font=\small\bfseries] {TPB $\rightarrow$ HA-CoPB};

\node[label, right=0.15cm of att.east, anchor=west, text width=3cm]
    {Individual intrinsic utility};
\node[label, right=0.15cm of sn.east, anchor=west, text width=3cm]
    {Intra-household role obligations};
\node[label, right=0.15cm of pbc.east, anchor=west, text width=3cm]
    {Shared resource constraints};

\node[below=1.5cm of pbc, font=\small\bfseries, align=center]
    (hier) {HA-CoPB Priority:\\
    $\mathcal{PBC} \succ \mathcal{SN} \succ \mathcal{A}$};

\end{tikzpicture}
\caption{The Theory of Planned Behavior (TPB) and the proposed HA-CoPB
adaptation. Our hierarchy mandates that shared resource barriers
($\mathcal{PBC}$) override social obligations ($\mathcal{SN}$),
which in turn override individual preferences ($\mathcal{A}$).}
\label{fig:tpb_diagram}
\end{figure*}

\subsection{Translation Map and Neutralization Filter Details}
To ensure the persona narratives are factually grounded and free of dataset-specific structural leakage, we map raw survey codes to deterministic natural language statements. Table \ref{tab:translation_map_travel} provides the mapping logic for the travel domain, while Table \ref{tab:translation_map_psid} provides the longitudinal mapping logic for the PSID residential mobility domain.

\begin{table*}[h]
\centering
\small
\begin{tabular}{lp{4cm}p{7cm}}
\toprule
\textbf{Variable} & \textbf{Raw Code Example} & \textbf{Natural Language Mapping} \\
\midrule
\textit{Demographics} & & \\
\texttt{R\_AGE\_IMP} & $34$ & "I am 34 years old." \\
\texttt{R\_SEX\_IMP} & $1$ (Male), $2$ (Female) & "I am a man/woman." (uses "boy/girl" if Age $<18$) \\
\texttt{EDUC} & $5$ (Grad Degree) & "I have a graduate or professional degree." \\
\texttt{HHFAMINC} & $11$ ($>\$200k$) & "My household income is \$200,000 or more." \\
\midrule
\textit{Household} & & \\
\texttt{R\_RELAT} & $3$ (Child) & "I am the son/daughter of the household head." \\
\texttt{HHVEHCNT} & $0$ & "My household does not own any vehicles." \\
\texttt{LIF\_CYC} & $4$ (2+ Adults, Young Child) & "My household consists of 2 adults, 1 child, youngest child aged 0-5." \\
\midrule
\textit{Context} & & \\
\texttt{URBRUR} & $1$ (Urban) & "I live in an urban area." \\
\texttt{RAIL} & $1$ (Has Rail) & "My metropolitan area has heavy rail." \\
\texttt{PHYACT} & $3$ (Vigorous) & "I engage in some vigorous physical activities." \\
\bottomrule
\end{tabular}
\caption{Travel Domain Variable Translation Logic. Raw codes from the travel survey are deterministically mapped to natural language atoms before narrative synthesis.}
\label{tab:translation_map_travel}
\end{table*}

\begin{table*}[h]
\centering
\small

\begin{tabular}{lp{4cm}p{7cm}}
\toprule
\textbf{Variable} & \textbf{Raw Code Example} & \textbf{Natural Language Mapping} \\
\midrule
\textit{Demographics} & & \\
\texttt{AGE} & $45$ & "I am a 45-year-old man/woman." \\
\texttt{EMPLOYMENT\_STATUS} & $2$ (Temporarily laid off) & "Regarding employment, I am temporarily laid off." \\
\texttt{DEGREE\_TYPE} & $2$ (Bachelor's) & "My highest degree is a bachelor's degree." \\
\midrule
\textit{Household} & & \\
\texttt{RELATION\_TO\_HEAD} & $10$ (Reference Person) & "I am the reference person (head) of this household." \\
\texttt{income\_to\_needs} & $3.5$ & "Household context: family income relative to census needs standard is about 3.50." \\
\texttt{HHVEHCNT} & $2$ & "Household context: the household owns 2 vehicle(s)." \\
\midrule
\textit{Longitudinal Context} & & \\
\texttt{bh\_tenure\_years\_2021} & $4.0$ (Years in home) & "Prior-wave snapshot: the household's tenure at the current residence AS OF MID-2021 was at least 4 years." \\
\texttt{bh\_move\_count\_prior} & $2$ (Moves in prior waves) & "Panel history: the household moved twice across those two prior waves." \\
\bottomrule
\end{tabular}
\caption{Residential Mobility (PSID) Variable Translation Logic. To avoid leaking the move-outcome, we explicitly enforce temporal labeling (e.g., "Prior-wave snapshot") and neutralize relationship categories.}
\label{tab:translation_map_psid}
\end{table*}

\subsection{Behavioral Anchor Logic}
\label{app:anchors}
Consistent with the principle of temporal model transferability \cite{atherton1976transferability}, the model treats the historical anchor as a prompt-based prior. 
\begin{itemize}
    \item \textbf{Continuous Formulation (Travel):} The prior $y_{\text{prior}}^{t-\text{lag}}$ is represented as a continuous average frequency (e.g., mean daily trip rate $\lambda$) calculated across historical demographic sub-groups based on primary identifiers (e.g., age bracket, vehicle ownership, urban/rural status). 
    \item \textbf{Probability Formulation (Mobility):} The prior $y_{\text{prior}}^{t-\text{lag}}$ is represented as a baseline relocation probability synthesized through multi-factor weighted blending of historical sub-populations. It serves as an expected base rate which the AvR Moderator evaluates against qualitative deviations in the agent's HA-CoPB reasoning.
\end{itemize}

\subsection{Attitudinal Markers and Imputation Rules}
\label{app:attitudes}
The Attitude ($\mathcal{A}$) construct provides the intrinsic psychological drivers for the persona. Table \ref{tab:attitudinal_markers} details the specific latent psychographic markers inferred for both the Travel and Mobility domains, alongside their specific HA-CoPB implications.

\begin{table*}[h]
\centering
\small
\begin{tabular}{p{2cm}p{2.5cm}p{4cm}p{5.5cm}}
\toprule
\textbf{Domain} & \textbf{Construct} & \textbf{Narrative Statement} & \textbf{HA-CoPB Implication} \\
\midrule
Travel & Tech-Savvy & "Learning new technologies is often frustrating for me. (Reverse)" & High PBC for app-based mobility and e-shopping substitution. \\
Travel & Polychronic & "I prefer to do one thing at a time. (Reverse)" & Tolerance for longer transit/ride-hail trips if productive; justifies complex trip chaining. \\
Travel & Materialistic & "I enjoy having a lot of luxury things." & Vehicle Preference: Correlates with private vehicle usage and higher discretionary shopping generation. \\
Travel & Pro-Car & "I like the idea of driving as a means of travel." & Resistance to mode switching; high utility for driving. \\
Travel & Pro-Transit & "I like the idea of public transit as a means of travel." & Increases probability of transit usage if PBC allows. \\
Travel & Wait Tolerant & "The trip itself is a waste of time. (Reverse)" & Lower disutility of travel time; longer commute acceptance. \\
\midrule
Mobility (PSID) & Place-Attached & "I prefer to stay in my current neighborhood." & Lower baseline mobility; tenure and local ties increase moving costs in the PBC layer. \\
Mobility (PSID) & Career-Driven & "I am willing to relocate to advance my career." & Job-linked mobility; strengthens move intentions when employment shocks appear. \\
Mobility (PSID) & Family-Oriented & "Schooling and space needs come first." & Crowding increases subjective-norm pressure toward larger housing. \\
Mobility (PSID) & Cost-Constrained & "Housing cost burdens make it hard to move." & PBC frictions: affordability limits feasible moves; may force distress moves. \\
Mobility (PSID) & Lifestyle-Seeker & "I value flexibility and am open to new cities." & Higher discretionary mobility among renters and young adults. \\
Mobility (PSID) & Stability-Seeker & "I prioritize predictable housing and avoiding disruption." & Strong inertia; homeownership and lifecycle stability reduce realized mobility. \\
\bottomrule
\end{tabular}
\caption{Attitudinal Constructs mapped to their HA-CoPB theoretical implications across both domains.}
\label{tab:attitudinal_markers}
\end{table*}

\subsection{Hierarchical Priority \& Construct Logic}
\label{app:priority_logic}
To prevent LLM hallucination and ground predictions in behavioral realism, the generation process strictly enforces the Hierarchical Priority Mechanism ($\mathcal{PBC} \succ \mathcal{SN} \succ \mathcal{A}$).
\begin{itemize}
    \item \textbf{Travel Domain Constraints:} An agent must evaluate hard resource limitations ($\mathcal{PBC}$) before considering normative obligations ($\mathcal{SN}$) or intrinsic desires ($\mathcal{A}$). For instance, if a household has 0 vehicles (PBC), a "Pro-Car" attitude (A) cannot trigger a private vehicle trip. Similarly, if there is a vehicle bottleneck (e.g., 2 drivers, 1 car), the model must simulate negotiation rather than hallucinating independent parallel trips.
    \item \textbf{Mobility Domain Constraints (PSID):} Structural immobility ($\mathcal{PBC}$) dominates desire-driven relocation ($\mathcal{A}$). If a household owns their home (a massive transaction friction) or suffers from acute labor shocks with no wealth cushion, the PBC acts as a hard veto against moving. Conversely, deep poverty often traps households structurally, overriding a desire for better neighborhoods. Only when PBC constraints are satisfied can the Subjective Norms (e.g., family crowding) and Attitudes (e.g., Career-Driven) dictate the final AvR consensus.
\end{itemize}

\section{Revision Clamping in Consensus Refinement}
\label{app: clamping}
During the consensus refinement process, we did revision clamping for each household member's utterance at each round for numerical decisions (with binary decisions unchanged). Specifically, the per-round decision update is limited to at most a fraction $\rho_{\text{turn}}$ of the speaker’s previous value. In addition, the preferred decision is bounded relative to the aggregated initial proposal:
\begin{equation}
\hat{y} \in \left[(1-\rho_{\text{down}})\hat{y}^{(0)}, (1+\rho_{\text{up}})\hat{y}^{(0)}\right],
\end{equation}
where $\hat{y}$ denotes the current preferred decision, and $\rho_{\text{down}}$ and $\rho_{\text{up}}$ control the allowable downward and upward deviation from the initial household proposal. This bound prevents the refined decision from drifting excessively from members’ initial intentions while still allowing reasonable updates during consensus refinement.

\section{Perception Validation Details}
\label{app:validation}

\paragraph{Survey Instruments}
To rigorously evaluate whether the synthesized personas possess a coherent worldview consistent with their demographics, we validate their responses against specific attitudinal variables held out from the generative phase. For the Travel Domain (NHTS), we preserve the exact wording and Likert scales used in the original survey:

\begin{itemize}
    \item \textbf{Health Status (HEALTH):} \\
    \textit{Question:} "In general, would you say your health is?" \\
    \textit{Scale:} 1=Excellent, 2=Very Good, 3=Good, 4=Fair, 5=Poor.
    
    \item \textbf{Price Sensitivity (PRICE):} \\
    \textit{Statement:} "The price of gasoline affects the number of vehicle trips I make." \\
    \textit{Scale:} 1=Strongly Agree, 2=Agree, 3=Neither Agree or Disagree, 4=Disagree, 5=Strongly Disagree.
    
    \item \textbf{Walkability Preference (PLACE):} \\
    \textit{Statement:} "I prefer to live in a community with mixed land uses (homes, shops, work) so I can walk to places." \\
    \textit{Scale:} 1=Strongly Agree, 2=Agree, 3=Neither Agree or Disagree, 4=Disagree, 5=Strongly Disagree.
\end{itemize}

For the Mobility Domain (PSID), we leverage longitudinal well-being and distress indicators:

\begin{itemize}
    \item \textbf{Food Affordability (FOOD\_AFFORD):} \\
    \textit{Statement:} "We couldn't afford to eat balanced meals." \\
    \textit{Scale:} 1=Often True, 2=Sometimes True, 3=Never True.

    \item \textbf{Life Satisfaction (LIFE\_SAT):} \\
    \textit{Question:} "Please think about your life-as-a-whole. How satisfied are you with it?" \\
    \textit{Scale:} 1=Completely Satisfied to 5=Not at all Satisfied.

    \item \textbf{Psychological Distress (K6\_DISTRESS):} \\
    \textit{Measure:} Aggregate K6 distress scale mapped to ordinal quintiles. \\
    \textit{Scale:} 1=Low Distress to 5=High Distress.
\end{itemize}

\paragraph{Tabular Results}
Figure~\ref{fig:perception_validation} in the main text visualizes the diagonal-split correlation heatmaps underlying the aggregate alignment scores below. Table~\ref{tab:app_perception_results} provides the detailed quantitative breakdown of perception validation performance across all evaluated metrics for both datasets.

\begin{table*}[ht]
\centering
\small
\begin{tabular}{lccc|ccc}
\toprule
& \multicolumn{3}{c|}{\textbf{Travel Domain (NHTS)}} & \multicolumn{3}{c}{\textbf{Mobility Domain (PSID)}} \\
\cmidrule(lr){2-4} \cmidrule(lr){5-7}
\textbf{Metric} & \textbf{Health} & \textbf{Price} & \textbf{Place} & \textbf{Food Afford.} & \textbf{Life Sat.} & \textbf{K6 Distress} \\
\midrule
Accuracy (Exact) & 38.1\% & 26.2\% & 27.7\% & 71.0\% & 41.4\% & 47.9\% \\
\textbf{Accuracy ($\pm 1$)} & \textbf{85.0\%} & \textbf{64.3\%} & \textbf{73.4\%} & \textbf{94.7\%} & \textbf{88.4\%} & \textbf{87.7\%} \\
MAE & 0.80 & 1.32 & 1.07 & 0.34 & 0.71 & 0.69 \\
\midrule
QWK (Kappa) & 0.29 & 0.02 & 0.03 & 0.11 & 0.18 & 0.18 \\
Wasserstein ($W_1$) & 0.47 & 0.86 & 0.28 & 0.12 & 0.26 & 0.33 \\
Spearman ($r_s$) & 0.29 & 0.01 & 0.03 & 0.15 & 0.18 & 0.19 \\
\midrule
\textbf{Aggregate Structural Align ($\rho$)} & \multicolumn{3}{c|}{\textbf{0.70}} & \multicolumn{3}{c}{\textbf{0.82}} \\
\bottomrule
\end{tabular}

\caption{\textbf{Comprehensive Perception Validation Results.} Performance of PEMAND across all targeted perception probes in both domains.}
\label{tab:app_perception_results}
\end{table*}

\begin{figure*}[ht]
\centering
\includegraphics[width=\textwidth]{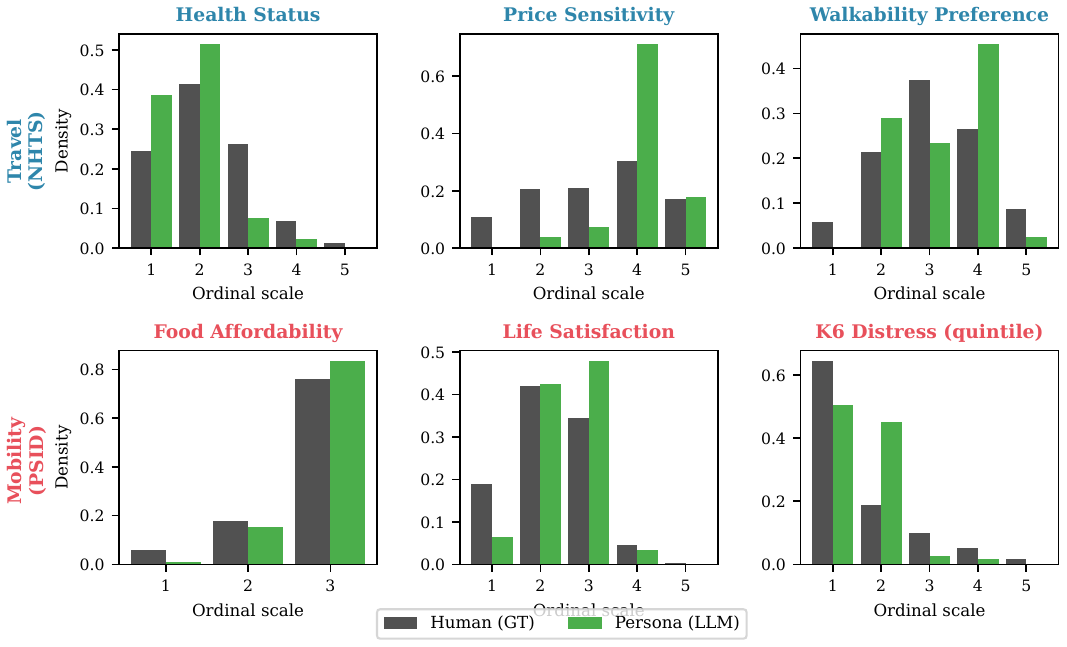}
\caption{\textbf{Perception probe distributional fidelity.} Normalized histograms comparing human ground truth (dark) vs.\ synthesized persona responses (green) for all six holdout probes. Top row: Travel domain (NHTS). Bottom row: Mobility domain (PSID).}
\label{fig:app_perception_histograms}
\end{figure*}

\paragraph{Distributional Histograms}
Figure~\ref{fig:app_perception_histograms} compares the marginal response distributions for all six holdout probes. Each panel overlays the human ground-truth density (dark) against the synthesized persona density (green). K6 distress is shown on quintile-binned ordinal scores (1=low to 5=high), consistent with the individual-level metrics in Table~\ref{tab:app_perception_results}. Visual inspection confirms that personas track the dominant modes for well-structured probes (e.g., \textit{Health Status}, \textit{Food Affordability}) while revealing expected smoothing on highly subjective items (e.g., \textit{Price Sensitivity}).

\paragraph{Mathematical Formulations of Metrics}

\subparagraph{Accuracy within a Tolerance ($\pm 1$ Scale Point)}
We adapt tolerance-based accuracy to Likert-scale prediction by treating a response as accurate when $|y_{i,v} - \hat{y}_{i,v}| \leq 1$. This $\pm 1$ tolerance reflects the subjectivity of attitude ratings (e.g., the subtle distinction between "Agree" and "Strongly Agree") and captures whether the persona’s sentiment is directionally correct.

\subparagraph{Quadratic Weighted Kappa (QWK)}
To measure individual-level agreement while accounting for the ordinal nature of Likert scales, we use Quadratic Weighted Kappa \cite{cohen1968weighted}. Unlike simple accuracy, QWK penalizes large disagreements (e.g., predicting 1 vs 5) significantly more than minor shifts (e.g., 1 vs 2):
\begin{equation}
\kappa_w = 1 - \frac{\sum_{i,j} w_{ij} O_{ij}}{\sum_{i,j} w_{ij} E_{ij}}
\end{equation}
where $w_{ij} = (i-j)^2$ is the quadratic weight, $O_{ij}$ is the observed confusion matrix, and $E_{ij}$ is the expected matrix under independence.

\subparagraph{Wasserstein Distance (EMD)}
To ensure the population of agents reproduces the aggregate public opinion distribution rather than collapsing to the mode, we calculate the Wasserstein Distance (Earth Mover's Distance) \cite{santurkar2023whose} between the cumulative distribution functions (CDF) of the human ($U$) and agent ($V$) responses:
\begin{equation}
W_1(U, V) = \sum_{k=1}^{K} |CDF_U(k) - CDF_V(k)|
\end{equation}
where $K$ is the number of ordinal Likert scale points. A lower Wasserstein distance indicates that the shape of the simulated opinion distribution closely matches the human ground truth.

\subparagraph{Structural Correlation Alignment}
To verify that the model captures latent sociological mechanisms, we compute the \textit{Structural Correlation Alignment} \cite{aher2023using}. For each domain, we form a correlation vector from all Spearman correlations between demographics and held-out probe responses. For NHTS, demographics are age, household income, tract population density, and vehicle count; for PSID, they are reference-person age, family income, home-ownership status, and self-rated health of the household head (\texttt{RP\_HEALTH}). Each domain therefore contributes twelve demographic--probe pairs ($4 \times 3$). Urbanicity was unavailable in our 2023 PSID extract; we include health because it appears in persona facts but is excluded from the held-out wellbeing probes. Let $\rho_{obs}$ denote the human vector and $\rho_{sim}$ the persona vector. The alignment score is:
\begin{equation}
\text{Alignment} = \text{Spearman}(\rho_{obs}, \rho_{sim})
\end{equation}
A high positive value indicates that the agents replicate the underlying relationships (e.g., higher income associating with lower food-affordability stress) found in the real world.

\section{Moderator Human Calibration Details}
\label{app:moderator-calibration}

To ensure that the LLM judge reliably enforces constraints during intra-conversation control, we conducted a human calibration study. We recruited four graduate student annotators from different countries and with diverse gender backgrounds, all of whom had expertise in human decision-making. The annotators participated voluntarily and received no monetary compensation.

\subsection{Annotation Rubric and Protocol}
A panel of four human annotators evaluated the model outputs double-blind on a 1--3 scale (1=Poor, 2=Adequate, 3=Good) across four dimensions:
\begin{enumerate}
    \item \textbf{Persona Consistency (per-member):} Evaluates whether the agent stays in character and reflects their socioeconomic profile without hallucinating false identities.
    \item \textbf{Constraint Adherence (per-member):} Evaluates if the agents perfectly obey the physical, financial, and temporal limits of the household.
    \item \textbf{Interaction Validity (holistic):} Evaluates the quality, flow, and realism of the multi-agent negotiation (e.g., dynamic deliberation vs. repetitive loops).
    \item \textbf{Decision Quality (holistic):} Evaluates the logical soundness and mathematical accuracy of the final agreed-upon outcome.
\end{enumerate}

Annotations were conducted in batches with a 20\% cross-annotator overlap to monitor drift. Inter-Rater Reliability (IRR) was computed continuously to ensure alignment.

\subsection{Inter-Rater Reliability (IRR) and Alignment Results}

Table~\ref{tab:irr-calibration} reports the multi-rater agreement (Fleiss' $\kappa$ \cite{fleiss1971measuring}) and the average pairwise Quadratic Weighted Kappa (QWK \cite{cohen1968weighted}) for the human annotators. Additionally, to demonstrate the successful transfer of these human preferences to the LLM judge, we compare the LLM--Human QWK before and after role-conditioned SFT on the Travel (Puget Sound) domain.

\begin{table*}[!ht]
\centering
\small
\begin{tabular}{lcccccc}
\toprule
& \multicolumn{3}{c}{\textbf{Travel (Puget Sound)}} & \multicolumn{3}{c}{\textbf{Mobility (PSID)}} \\
\cmidrule(lr){2-4} \cmidrule(lr){5-7}
\textbf{Dimension} & \textbf{Human} & \textbf{LLM (Before)} & \textbf{LLM (After)} & \textbf{Human} & \textbf{LLM (Before)} & \textbf{LLM (After)} \\
& \textbf{QWK} & \textbf{QWK} & \textbf{QWK} & \textbf{QWK} & \textbf{QWK} & \textbf{QWK} \\
\midrule
Persona Consistency & 0.446 & 0.133 & 0.450 & 0.847 & 0.139 & 0.490 \\
Constraint Adherence & 0.626 & 0.188 & 0.533 & 0.517 & 0.162 & 0.471 \\
Interaction Validity & 0.491 & 0.272 & 0.527 & 0.500 & 0.236 & 0.538 \\
Decision Quality & 0.845 & 0.506 & 0.749 & 0.729 & 0.460 & 0.648 \\
\midrule
\textbf{Overall Average} & \textbf{0.698} & \textbf{0.275} & \textbf{0.567} & \textbf{0.851} & \textbf{0.249} & \textbf{0.536} \\
\bottomrule
\multicolumn{7}{l}{\textit{Note: Overall human Fleiss' $\kappa = 0.620$ (Travel) and $0.853$ (Mobility).}}
\end{tabular}
\caption{\textbf{Moderator calibration inter-rater reliability.} Average pairwise QWK by rubric dimension. Travel (Puget Sound) includes LLM--Human QWK before and after role-conditioned SFT; Mobility (PSID) reports human agreement only.}
\label{tab:irr-calibration}
\end{table*}

\begin{table*}[t]
\centering
\small
\begin{tabular}{ll}
\hline
\textbf{Variable} & \textbf{Description} \\
\hline
HHVEHCNT* & Number of vehicles in household \\
HHFAMINC* & Household income category \\
HOMEOWN* & Home ownership status \\
URBRUR & Urban vs rural household location \\
RAIL & Rail availability in household area \\
TRAVDAY & Travel day type (weekday vs weekend) \\
CHILDREN & Presence of children in household (under 18) \\
DRVRCNT & Number of licensed drivers in household \\
GASPRICE & Gas price on the travel day (cents) \\
GT1JBLWK & Indicator for having more than one job \\
HHSIZE* & Household size (number of members) \\
HH\_HISP & Hispanic status of household respondent \\
HTEEMPDN & Employment density cluster (workers per square mile) \\
HTHTNRNT & Renter-occupied housing percentage cluster \\
HTPPOPDN & Population density cluster (people per square mile) \\
HTRESDN & Housing unit density cluster (units per square mile) \\
LPACT & Frequency of light/moderate physical activity (past week) \\
MSASIZE & Metropolitan Statistical Area (MSA) population size category \\
PC & Frequency of desktop/laptop internet use \\
PHYACT & Physical activity level (ordinal) \\
SPHONE & Frequency of smartphone internet use \\
YOUNGCHILD* & Presence of young child in household (typically age 0--5) \\
MSACAT\_01 & MSA $\geq$ 1M population, with rail \\
MSACAT\_02 & MSA $\geq$ 1M population, without rail \\
MSACAT\_03 & MSA $<$ 1M population \\
MSACAT\_04 & Not in an MSA \\
LIF\_CYC\_02 & Household life cycle: 2+ adults, no children \\
LIF\_CYC\_03 & Household life cycle: one adult, youngest child age 0--5 \\
LIF\_CYC\_04 & Household life cycle: 2+ adults, youngest child age 0--5 \\
LIF\_CYC\_05 & Household life cycle: one adult, youngest child age 6--15 \\
LIF\_CYC\_06 & Household life cycle: 2+ adults, youngest child age 6--15 \\
LIF\_CYC\_07 & Household life cycle: one adult, youngest child age 16--21 \\
LIF\_CYC\_08 & Household life cycle: 2+ adults, youngest child age 16--21 \\
LIF\_CYC\_09 & Household life cycle: one adult, retired, no children \\
LIF\_CYC\_10 & Household life cycle: 2+ adults, retired, no children \\
AGE\_U18\_PCT* & Percent of household members under age 18 \\
AGE\_18\_35\_PCT* & Percent of household members age 18--35 \\
AGE\_35\_54\_PCT* & Percent of household members age 35--54 \\
AGE\_55\_64\_PCT* & Percent of household members age 55--64 \\
AGE\_65P\_PCT* & Percent of household members age 65 and older \\
MALE\_PCT* & Percent of household members who are male \\
WHITE\_PCT* & Percent of household members who are White \\
MEDCOND\_PCT & Percent of household members with medical conditions \\
WORKER\_PCT* & Percent of household members who are employed \\
DRIVER\_PCT* & Percent of household members who are licensed drivers \\
OCCAT\_01\_SalesService\_PCT & Percent of household members in Sales/Service occupations \\
OCCAT\_02\_ClericalAdmin\_PCT & Percent of household members in Clerical/Administrative occupations \\
OCCAT\_03\_ManualLabor\_PCT & Percent of household members in Manual/Farming/Labor occupations \\
OCCAT\_04\_Professional\_PCT & Percent of household members in Professional/Technical occupations \\
OCCAT\_97\_Other\_PCT & Percent of household members in Other occupation categories \\
EDUC\_OTHER\_PCT* & Percent of household members with education below bachelor (or other non-degree category) \\
EDUC\_BACHELOR\_PCT* & Percent of household members with a bachelor degree \\
EDUC\_GRAD\_PCT* & Percent of household members with a graduate degree \\
STUDENT\_PCT* & Percent of household members who are students \\
GT1JBLWK\_PCT* & Percent of household members who have more than one job \\
POOR\_HEALTH\_PCT & Percent of household members reporting poor health \\
LPACT\_MEAN & Mean light/moderate activity frequency across household members \\
PHYACT\_LIGHT\_PCT & Percent of household members with light activity level \\
PHYACT\_MODERATE\_PCT & Percent of household members with moderate activity level \\
PHYACT\_VIGOROUS\_PCT & Percent of household members with vigorous activity level \\
RIDESHARE\_MEAN & Mean rideshare usage count across household members \\
PTUSED\_MEAN & Mean public transit usage count across household members \\
BORNINUS\_PCT & Percent of household members born in the United States \\
\hline
\end{tabular}
\caption{Variable names and descriptions for NHTS 2017. Variables marked with ``*'' appear in both NHTS and Puget Sound.}
\label{tab:variables}
\end{table*}

\begin{table*}[t]
\centering
\small
\begin{tabular}{ll}
\hline
\textbf{Variable} & \textbf{Description} \\
\hline
HHSIZE & Household size \\
NUM\_CHILDREN\_FU & Number of children in the family unit \\
AGE\_YOUNGEST\_CHILD & Age of the youngest child in the household \\
MARITAL\_STATUS* & Marital status of the reference person \\
HOMEOWN* & Home ownership or housing tenure status \\
STATE\_FIPS* & State location code \\
RACE\_RP* & Race of the reference person \\
ETHNIC\_RP* & Ethnic identification of the reference person \\
IMMIGRANT\_SAMPLE* & Indicator for PSID immigrant sample membership \\
RP\_AGE & Age of the reference person \\
SP\_AGE & Age of spouse or partner \\
RP\_SEX* & Sex of the reference person \\
SP\_SEX* & Sex of spouse or partner \\
RP\_HEALTH & Self-reported health of the reference person \\
SP\_HEALTH & Self-reported health of spouse or partner \\
RP\_EDUC\_YEARS & Years of education of the reference person \\
SP\_EDUC\_YEARS & Years of education of spouse or partner \\
HHVEHCNT & Number of vehicles in household \\
HHFAMINC & Total household/family income \\
NEEDS\_STANDARD\_2022 & Census needs standard for the household in 2022 \\
INCOME\_TO\_NEEDS & Household income-to-needs ratio \\
CREDIT\_CARD\_DEBT & Household credit card debt \\
WEALTH\_NO\_EQUITY & Household wealth excluding home equity \\
BIRTH\_RP\_SP\_2021 & Birth-related measure for reference person/spouse in 2021 \\
BIRTH\_2021\_FLAG* & Indicator for a birth in 2021 \\
RP\_WEEKS\_LAIDOFF\_2022 & Weeks reference person was laid off in 2022 \\
SP\_WEEKS\_LAIDOFF\_2022 & Weeks spouse/partner was laid off in 2022 \\
RP\_WEEKS\_UNEMP\_2022 & Weeks reference person was unemployed in 2022 \\
RP\_WEEKS\_UNEMP\_2021 & Weeks reference person was unemployed in 2021 \\
SP\_WEEKS\_UNEMP\_2021 & Weeks spouse/partner was unemployed in 2021 \\
RP\_LUNG\_LIMIT\_ACTIVITY & Reference person activity limitation due to lung disease \\
RP\_ARTHRITIS\_LIMIT\_ACTIVITY & Reference person activity limitation due to arthritis \\
SP\_ARTHRITIS\_LIMIT\_ACTIVITY & Spouse/partner activity limitation due to arthritis \\
ANY\_LAYOFF\_2022* & Indicator for any household layoff in 2022 \\
ANY\_UNEMP\_2022* & Indicator for any household unemployment in 2022 \\
ANY\_UNEMP\_2021* & Indicator for any household unemployment in 2021 \\
ANY\_DISABILITY\_LIMIT* & Indicator for any household disability/activity limitation \\
MAX\_YEARS\_EDUC & Maximum years of education in the household \\
MAX\_NUM\_JOBS\_PY & Maximum number of jobs held in the prior year \\
PERSON\_COUNT & Number of persons represented in household records \\
CHILDREN\_PCT & Share of household members who are children \\
MALE\_PCT & Share of household members who are male \\
SENIOR\_PCT & Share of household members who are seniors \\
AGE\_U18\_PCT & Share of household members under age 18 \\
AGE\_18\_35\_PCT & Share of household members aged 18--35 \\
AGE\_35\_54\_PCT & Share of household members aged 35--54 \\
AGE\_55\_64\_PCT & Share of household members aged 55--64 \\
AGE\_65P\_PCT & Share of household members aged 65 or older \\
EMPLOYED\_PCT & Share of household members employed \\
RETIRED\_PCT & Share of household members retired \\
STUDENT\_PCT & Share of household members who are students \\
DISABLED\_PCT & Share of household members with disability status \\
HEALTHY\_PCT & Share of household members reporting good health \\
MULTI\_JOB\_PCT & Share of household members holding multiple jobs \\
EDUC\_BACHELOR\_PCT & Share of household members with a bachelor's degree \\
EDUC\_GRAD\_PCT & Share of household members with graduate education \\
COLLEGE\_DEGREE\_PCT & Share of household members with a college degree \\
\hline
\end{tabular}
\caption{Processed PSID variables used for residential mobility prediction. Variables marked with * are treated as categorical variables in the preprocessing pipeline.}
\label{tab:psid_variables}
\end{table*}

\end{document}